\documentclass[10pt,twocolumn,letterpaper]{article}

\usepackage[pagenumbers]{cvpr} % To force page numbers, e.g. for an arXiv version
\usepackage{multirow}
\usepackage{xcolor}
\usepackage[table]{xcolor}

\definecolor{darkgreen}{RGB}{0,100,0}
\definecolor{cvprblue}{rgb}{0.21,0.49,0.74}
\usepackage[pagebackref,breaklinks,colorlinks,allcolors=cvprblue]{hyperref}
\usepackage[most]{tcolorbox} % 导言区添加

\definecolor{cvprblue}{rgb}{0.21,0.49,0.74}
\usepackage[pagebackref,breaklinks,colorlinks,allcolors=cvprblue]{hyperref}

\title{
Photo3D: Advancing Photorealistic 3D Generation through \\ Structure‑Aligned Detail Enhancement \\
\vspace{6pt}
\small\url{https://liangsanzhu.github.io/photo3d-page/}
}
\makeatletter
\newcommand\blfootnote[1]{%
  \begingroup
  \renewcommand\thefootnote{}% 去掉编号
  \footnotetext{#1}% 打脚注内容
  \addtocounter{footnote}{-1}% 不让计数器+1
  \endgroup
}
\makeatother
\author{
Xinyue Liang \quad
Zhiyuan Ma \quad
Lingchen Sun \quad
Yanjun Guo \quad
Lei Zhang\footnotemark\\[4pt]
Department of Computing, The Hong Kong Polytechnic University\\
Hong Kong, China\\
{\tt\small \{xinyue.liang, zm2354.ma, ling-chen.sun, yanjunn.guo\}@connect.polyu.hk, cslzhang@polyu.edu.hk}
}
\begin{document}
%\maketitle
\twocolumn[{%
\renewcommand\twocolumn[1][]{#1}%
\maketitle
\begin{center}

\includegraphics[width=1\textwidth]{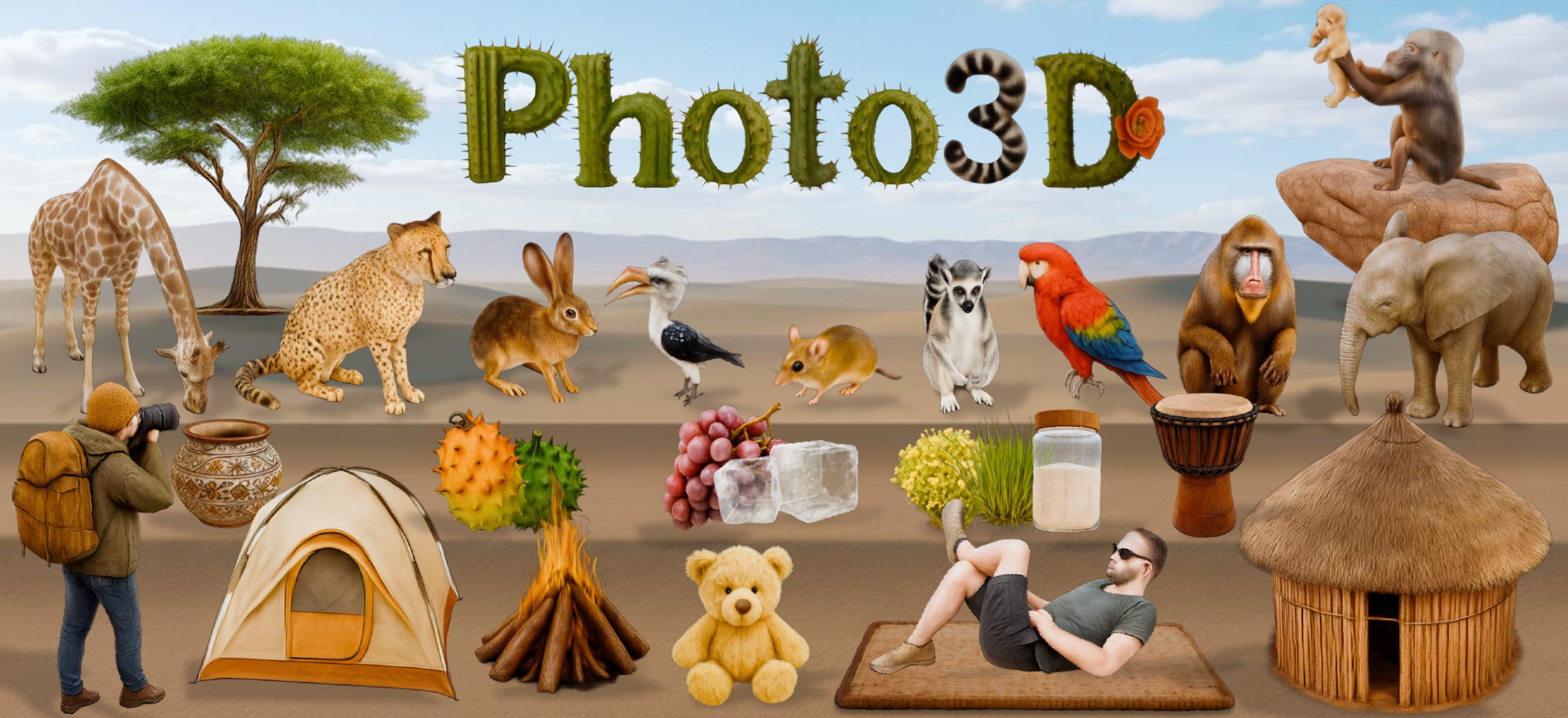}
\label{fig:teaser}
\end{center}

 \vspace{-2em}
\captionof{figure}{Photorealistic 3D objects in the foreground are generated by \textbf{Photo3D}, featuring rich realistic details and stable 3D structure.\vspace{1em}}

\label{fig:teaser}

}]
\blfootnote{* Corresponding author.}

\begin{abstract}
Although recent 3D‑native generators have made great progress in synthesizing reliable geometry, they still fall short in achieving realistic appearances. A key obstacle lies in the lack of diverse and high-quality real-world 3D assets with rich texture details, since capturing such data is intrinsically difficult due to the diverse scales of scenes, non‑rigid motions of objects, and the limited precision of 3D scanners.
We introduce \textbf{Photo3D}, a framework for advancing photorealistic 3D generation, which is driven by the image data generated by the GPT‑4o‑Image model.
Considering that the generated images can distort 3D structures due to their lack of multi‑view consistency, we design a structure‑aligned multi‑view synthesis pipeline and construct a detail‑enhanced multi‑view dataset paired with 3D geometry. 
Building on it, we present a realistic detail enhancement scheme that leverages perceptual feature adaptation and semantic structure matching to enforce appearance consistency with realistic details while preserving the structural consistency with the 3D-native geometry. Our scheme is general to different 3D-native generators, and we present dedicated training strategies to facilitate the optimization of geometry-texture coupled and decoupled 3D-native generation paradigms. Experiments demonstrate that Photo3D generalizes well across diverse 3D‑native generation paradigms and achieves state‑of‑the‑art photorealistic 3D generation performance. 
 %on multiple benchmarks. 
\end{abstract}    
\section{Introduction}

3D generation has progressed rapidly in recent years across both academia and industry~\cite{poole2022dreamfusion,liu2023zero,tang2024lgm,zhao2025hunyuan3d,ma2025progressive,wang2026one2scene,chen2025fast,wang2024open}. Recent works have increasingly shifted from score‑distillation‑based~\cite{poole2022dreamfusion,wang2023prolificdreamer,ma2024scaledreamer} and multi‑view‑based approaches~\cite{liu2023zero,tang2024lgm,long2024wonder3d,chen2024lara} to 3D‑native generation methods~\cite{li2025step1x,xiang2025structured,zhao2025hunyuan3d,ye2025hi3dgen,li2025triposg}, which directly learn 3D distributions from large‑scale 3D datasets~\cite{deitke2023objaversexl,deitke2023objaverse,chang2015shapenet,collins2022abo} to enable faster, more stable and fine‑grained geometry generation. Generally, 3D‑native generators can be categorized into geometry–texture coupled and decoupled paradigms: the former jointly learn geometry and texture, while the latter first synthesize geometry and then generate textures with another 3D‑native or multi‑view-based texturing model. The models in both categories are trained or finetuned on large‑scale 3D datasets. However, existing large-scale 3D datasets~\cite{deitke2023objaversexl,deitke2023objaverse,chang2015shapenet,collins2022abo} are predominantly composed of synthetic assets that differ from natural imagery, as high-quality real‑world 3D collections remain scarce due to the difficulties of capturing data from diverse-scale, deformable and dynamic natural objects. Although some small-scale static real‑world 3D datasets are available~\cite{brazil2023omni3d,downs2022google,dong2025digital,reizenstein21co3d}, constraints in current 3D representations and scanner precision often lead to over-smoothed appearance, lacking fine‑grained texture details.
Consequently, existing 3D‑native generators often produce 3D models with synthetic coloration and cartoon‑like textures, exposing a persistent gap between geometric plausibility and appearance realism.

Compared with 3D data, 2D imagery offers richer and more realistic appearance details. To enhance the realism of 3D generation, the recent work Real3D~\cite{Jiang2025ICCV} employs single‑view image supervision with cycle‑consistent losses, yet the absence of stereo constraints leads to unstable geometry and limited around-view realism. A natural extension is to adopt multi-view real-world imagery for supervision, such as the rendered views from real 3D datasets~\cite{dong2025digital,downs2022google,brazil2023omni3d} or captured photo collections~\cite{reizenstein21co3d,han2024mvimgnet2}, but these datasets are not well-suited for generalizable training due to their restricted category coverage and insufficient quality of appearance details.
Alternatively, recent image generators~\cite{hurst2024gpt,flux2024,labs2025flux1kontextflowmatching,comanici2025gemini} can synthesize realistic multi‑view images but lack intrinsic consistency, leading to texture inconsistency and structural drift. Additional multi-view generation finetuning~\cite{long2024wonder3d,qiu2024richdreamer,li2024era3d} on 3D datasets still fails to fully resolve these inconsistencies and instead biases the generated appearances toward synthetic 3D styles. Therefore, how to construct more realistic and consistent training data, along with robust training strategies to enhance realistic details, remains an open challenge.

To tackle this challenge, we introduce \textbf{Photo3D}, a framework for geometric structure‑aligned detail enhancement in photorealistic 3D generation, as showcased in Fig.~\ref{fig:teaser}.
Our approach first generates realistic multi‑views that are well aligned with 3D assets to construct multi‑view consistent realistic details.
Specifically, we first create 3D assets via 3D-native generation methods to represent real‑world objects, which, however, may lack detailed appearances.
We then leverage the advanced image generator, GPT‑4o‑Image~\cite{hurst2024gpt}, to enhance the realistic appearance of 3D renderings by refining them with fine‑grained, texture‑rich details. Unlike previous geometry‑conditioned image generation methods~\cite{huang2025mv,zhao2025hunyuan3d, bensadoun2024meta}, our appearance‑anchored refinement can achieve perceptually finer alignment with the 3D structures, yielding photorealistic multi‑view images paired with their 3D geometry, which can provide high‑fidelity training data for detail-rich 3D‑native generation.

However, the newly introduced realistic details may vary slightly across views due to the inherent generative diversity of the image generator. Therefore, instead of enforcing strict pixel‑wise supervision, we propose a relaxed detail‑enhancement scheme to facilitate feature adaptation and semantic structure matching, allowing the model to learn realistic appearance details from visually consistent correspondences.
Moreover, since different 3D‑native generators adopt distinct generative paradigms and training pipelines, to effectively incorporate realism priors in training, we design additional training strategies for different paradigms.
In summary, our contributions are threefold:
\begin{itemize}
\item We propose \textbf{Photo3D}, a framework that aims for photorealistic 3D generation by improving realistic appearance details while preserving structural consistency. Dedicated strategies are presented for different 3D‑native paradigms to facilitate training.
\item We develop a 3D‑aligned multi‑view synthesis pipeline and construct a realism‑enhanced dataset, namely \textbf{Photo3D‑MV}, to support the training of photorealistic 3D-native generation models.
\item Extensive experiments demonstrate that Photo3D achieves \textbf{state‑of‑the‑art} photorealistic 3D generation across diverse 3D‑native paradigms and benchmarks.
\end{itemize}

\label{sec:intro}

\section{Related Work}
\begin{figure*}[t]
  \centering
  
   \includegraphics[width=1\linewidth]{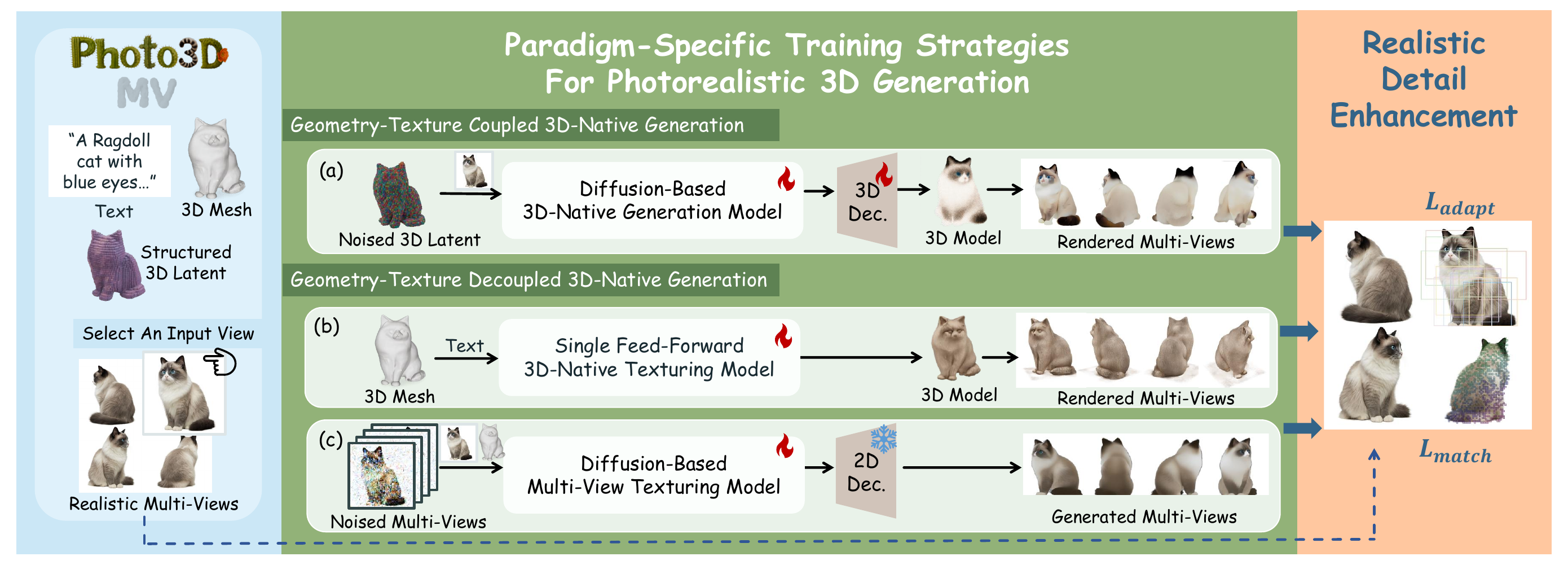}
\caption{\textbf{Overview of Photo3D.} We first construct Photo3D‑MV, a realistic, detail‑enhanced multi‑view dataset paired with 3D geometry, and propose associated schemes to learn realistic 3D appearance details.
Paradigm‑specific training strategies are designed for geometry–texture coupled and decoupled paradigms: (a) diffusion‑based 3D‑native generator (\eg, Trellis~\cite{xiang2025structured}); (b) single feed‑forward 3D‑native texturing model (\eg, TexGaussian~\cite{xiong2025texgaussian}); and (c) diffusion‑based multi‑view texturing model (\eg, Step1X-3D~\cite{li2025step1x}).}
\label{fig:pipeline}
%\vspace{-10pt} 
\end{figure*}

\noindent\textbf{3D Generation Paradigms}.
Recent 3D generation approaches can be divided into optimization‑based approaches~\cite{poole2022dreamfusion,wang2023prolificdreamer,chen2023fantasia3d,chen2024text,melas2023realfusion,tang2023make,ma2024scaledreamer}, which generate 3D models via score distillation but suffer from slow optimization, and learning‑based paradigms, which can be further categorized into 2D‑lifting and 3D‑native approaches. 2D‑lifting methods~\cite{liu2023zero,long2024wonder3d,li2024era3d,tang2024lgm,hong2023lrm,liang2025aligncvc} first synthesize multi‑view images and then reconstruct them into a 3D model, but often exhibit multi‑view inconsistency and limited geometric fidelity due to a lack of explicit 3D priors. In contrast, 3D‑native generation~\cite{xiang2025structured,zhao2025hunyuan3d,ye2025hi3dgen,li2025triposg,guo2025hyper3d,yushi2025gaussiananything,siddiqui2024meshgpt,zhao2025deepmesh,wang2024llama} directly learns 3D data distributions from large‑scale 3D datasets~\cite{deitke2023objaverse,deitke2023objaversexl,chang2015shapenet,collins2022abo}, achieving more stable and fine‑grained geometry.

\vspace{+2mm}
\noindent\textbf{3D-Native Generation Pipelines}.
3D‑native generation methods can be categorized into geometry–texture coupled and decoupled approaches.
Geometry–texture coupled approaches~\cite{chen20253dtopia,xiang2025structured,yushi2025gaussiananything,lin2025diffsplat} jointly learn geometry and texture within a unified 3D framework, enabling end‑to‑end generation of consistent 3D geometry and texture.
Decoupled approaches~\cite{zhao2025hunyuan3d,ye2025hi3dgen,li2025triposg,li2025step1x,guo2025hyper3d,zhang2025bang} first generate 3D geometry and then infer texture in a subsequent stage.
In the texturing stage, 3D‑native texturing models~\cite{yu2024texgen,xiong2025texgaussian,liu2024texoct} learn 3D consistent appearance directly from 3D data, whereas multi‑view texturing models~\cite{huang2025mv,liang2025unitex,bensadoun2024meta,fei2025pacture,he2025materialmvp} use image generators finetuned on 3D datasets to synthesize textured multi-views, and then fuse them into UV texture maps.
These 3D‑native generation models rely heavily on 3D datasets~\cite{deitke2023objaversexl,deitke2023objaverse,chang2015shapenet,collins2022abo}, which are mainly composed of synthetic 3D assets, resulting in limited realistic details.

\vspace{+2mm}
\noindent\textbf{3D Photorealistic Enhancement}.
Recent efforts such as Real3D~\cite{Jiang2025ICCV} leverage single‑view real‑world images with cycle‑consistent supervision to improve appearance fidelity. However, the lack of multi‑view constraints causes geometric inconsistency and unstable cross‑view appearance.
Other approaches~\cite{he2025materialmvp,fei2025pacture,huang2025material,fang2024make} generate Physically Based Rendering (PBR) material maps but keep the original texture unchanged, adding only renderer‑dependent illumination and yielding limited improvements in appearance realism.
Meanwhile, real‑world scanned 3D datasets~\cite{downs2022google,dong2025digital,brazil2023omni3d} suffer from low diversity and insufficient texture detail due to limited scanning precision. %, reducing their utility for training 3D generators.
In contrast, our Photo3D enhances 3D appearance during the generative process, producing structure-aligned and rendering‑independent photorealistic 3D models that generalize across diverse 3D‑native generation frameworks.

\begin{figure*}[t]
  \centering
  
   \includegraphics[width=1\linewidth]{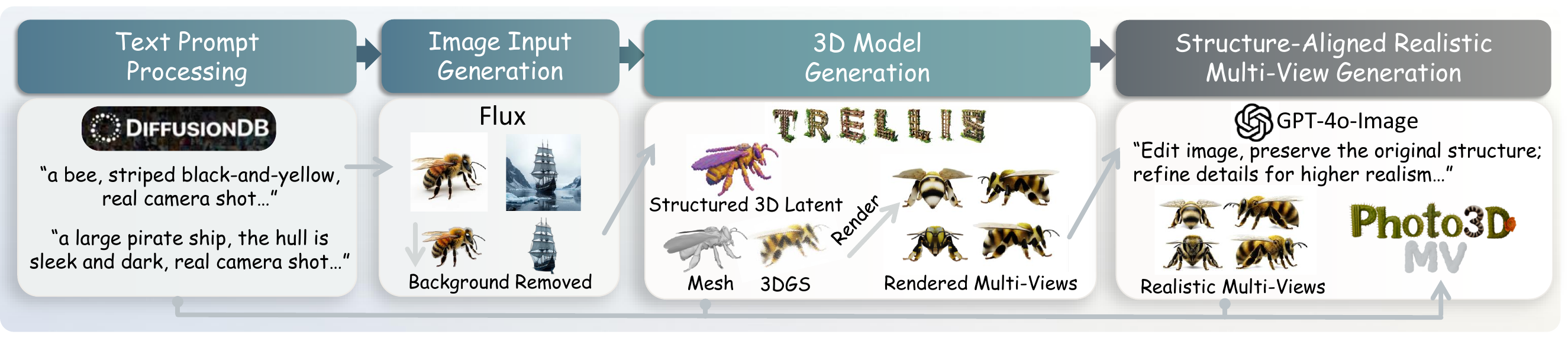}
\vspace{-7mm}
   \caption{The structure-aligned realistic multi-view synthesis pipeline for \textbf{Photo3D-MV} dataset. We first process text prompts from DiffusionDB~\cite{wang2022diffusiondb} to obtain object‑centric descriptions with realistic attributes. We then use Flux.1‑Dev~\cite{flux2024} to generate images, serving as inputs for 3D generation with Trellis~\cite{xiang2025structured}. Finally, we employ GPT‑4o‑Image~\cite{hurst2024gpt} to refine the multi‑view 3D renderings into structure‑aligned, photorealistic images. These realistic multi-views, together with text descriptions and the generated 3D assets, constitute Photo3D-MV.}
   \label{fig:dataset}
   \vspace{-2mm}
\end{figure*}

\begin{figure*}[t]
  \centering
      \vspace{-3mm}

  \includegraphics[width=0.9\linewidth]{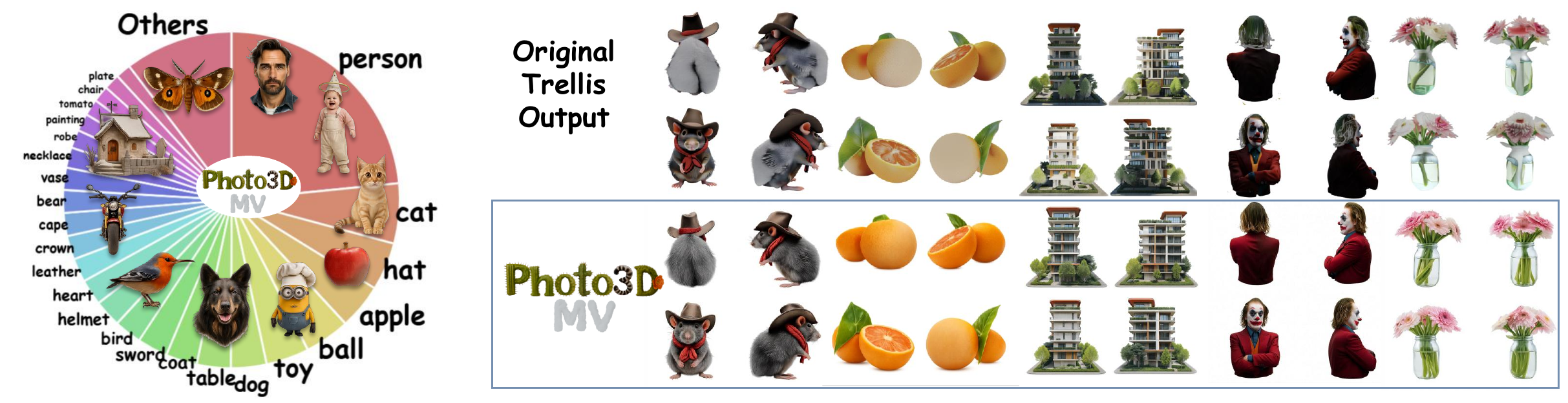}
    \vspace{-4mm}

\caption{Diverse distribution of top categories in Photo3D-MV.}
  \label{fig:bingtu}
  \vspace{-4mm}
\end{figure*}

\section{Method}
This section presents the proposed Photo3D method. We first develop a structure‑aligned multi‑view synthesis pipeline and construct a detail‑enhanced dataset, Photo3D‑MV (Sec.~\ref{sec:photo3dmv}). Building upon this, we propose a realistic detail enhancement scheme to refine appearance realism while preserving 3D structure fidelity (Sec.~\ref{sec:scheme}). Finally, we introduce paradigm‑specific training strategies to effectively integrate Photo3D into three representative 3D‑native generative paradigms (Sec.~\ref{sec:strategy}). An overview of the complete framework is shown in Fig.~\ref{fig:pipeline}.

\subsection{Photo3D-MV Dataset}
\label{sec:photo3dmv}
%Achieving photorealistic 3D generation remains challenging due to the limited availability of diverse, high‑quality real‑world 3D data. 
Capturing real‑world objects is challenging because they vary in scale and often exhibit non‑rigid motion, and existing scanned 3D assets usually lack fine‑grained surface details. To address this limitation, we leverage the advanced closed‑source image generator GPT‑4o‑Image~\cite{hurst2024gpt} to enhance the details of 3D renderings. While GPT‑4o‑Image can produce highly photorealistic images, its outputs often lack multi‑view consistency, leading to structural artifacts. Therefore, we introduce a structure‑aligned multi‑view synthesis pipeline and construct Photo3D‑MV, a detail‑enhanced multi‑view dataset paired with 3D geometry to provide realistic priors for 3D generation.

As illustrated in Fig.~\ref{fig:dataset}, we begin the multi‑view synthesis by processing text prompts from DiffusionDB~\cite{wang2022diffusiondb} with LLaMA‑3‑8B~\cite{llama3modelcard} to obtain object‑centric descriptions with enhanced photorealistic attributes. We then use Flux.1‑Dev~\cite{flux2024} to synthesize the corresponding single‑view images and remove their backgrounds. Next, we employ Trellis ~\cite{xiang2025structured}, a 3D-native generation model, to produce 3D assets, including structured 3D latents, 3D meshes, and 3D Gaussian Splatting (3DGS)~\cite{kerbl20233d} models. Subsequently, we render four orthogonal views of each 3DGS model, compose them into a single four‑panel image, and process this composite with GPT‑4o‑Image \cite{hurst2024gpt} using an editing prompt designed for cross‑view‑aligned detail enhancement. The refinement enriches the photorealistic appearance while preserving consistent 3D structure. Notably, compared with other image generators, such as Gemini‑2.5-Flash~\cite{comanici2025gemini} and Flux.1-Kontext~\cite{labs2025flux1kontextflowmatching}, GPT‑4o‑Image achieves superior fidelity in realistic detail enhancement. Finally, the resulting structure‑aligned multi‑view images and text descriptions, paired with the generated 3D meshes and 3D latents, constitute the Photo3D‑MV dataset, providing high-fidelity and 3D-aligned realistic detail priors. Photo3D‑MV contains 10K objects across 373 of the 1.2K LVIS~\cite{gupta2019lvis} categories, demonstrating diverse coverage, as shown in Fig.~\ref{fig:bingtu}.  More details of constructing Photo3D-MV are in the \textbf{Appendix~\ref{sec:Photo3D-MV}}.

\subsection{Realistic Detail Enhancement Scheme}
\label{sec:scheme}

While Photo3D‑MV provides strong realism priors with structure‑aligned multi‑view images of each 3D object, four multi-views are still insufficient to reconstruct fully textured 3D assets for 3D‑native training. Instead, we leverage these realistic images as supervision during training. However, GPT‑4o‑Image can introduce view‑dependent detail variations, and hence enforcing strict pixel‑level supervision can cause misalignment in structures and textures. To mitigate these issues, we propose a relaxed detail‑enhancement scheme to maintain structural stability.

\begin{figure}[t]
  \centering
  \includegraphics[width=1\linewidth]{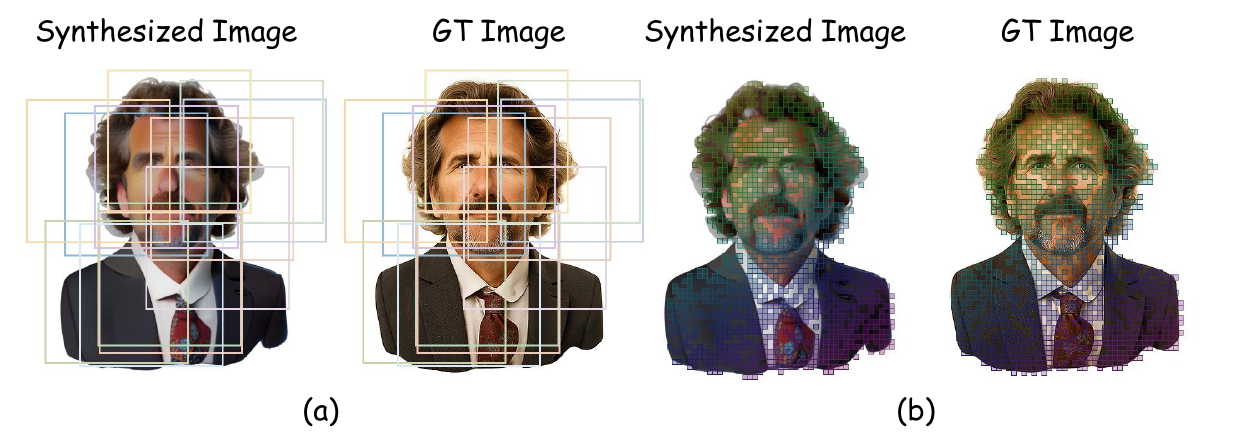}
  \vspace{-7mm}
\caption{
Computation of $\mathcal{L}_{\text{adapt}}$ and $\mathcal{L}_{\text{match}}$ between a synthesized image and the corresponding GT image.
(a) Same‑color boxes indicate crops for computing $\mathcal{L}_{\text{adapt}}$.
(b) Same‑color patches denote sampled semantically matched patches contributing to $\mathcal{L}_{\text{match}}$.
}
  \label{fig:loss}
  \vspace{-3mm}
\end{figure}

We first adopt perceptual feature adaptation to enhance the visual quality. %This adaptation is applied to synthesized images rendered from the generated textured 3D models or directly generated by the multi‑view texturing models, and the realistic images from Photo3D‑MV serve as their GT supervision.  However, direct supervision with an $\ell_2$ loss often leads to structural distortions. Although Gram loss~\cite{gatys2015texture,gatys2016image} and LPIPS loss~\cite{zhang2018unreasonable} can transfer some texture patterns and local feature details, they fail to preserve global structure consistency. 
Specifically, we adopt a CLIP-based loss~\cite{radford2021learning} that aligns synthesized and GT images within a shared embedding space, where the visual representations preserve high-level semantic information.  Since CLIP does not support high-resolution input, we apply random cropping to capture fine-grained details while preserving global perceptual coherence, as shown in Fig.~\ref{fig:loss}(a).
Formally, for a synthesized image $I_{\text{syn}}$ and the GT image $I_{\text{GT}}$, let $\tau_c(\cdot)$ denote a random crop, $\phi(\cdot)$ the CLIP encoder, and $\mathcal{C}$ the set of sampled crops. The perceptual feature adaptation loss is defined as:
\begin{equation}
\mathcal{L}_{\text{adapt}}
=
\frac{1}{|\mathcal{C}|}
\sum_{c\in\mathcal{C}}
\Big(
1 -
\langle
\phi(\tau_c(I_{\text{syn}})),
\phi(\tau_c(I_{\text{GT}}))
\rangle
\Big),
\label{eq:cliploss}
\end{equation}
where $\langle \cdot , \cdot \rangle$ denotes the cosine similarity in the CLIP embedding space. This random crop-wise formulation alleviates CLIP’s resolution limitation and enables the model to capture fine-grained details while maintaining perceptual coherence between the synthesized and GT images.

To further maintain structural stability, we introduce semantic structure matching,  which establishes correspondences between semantically related regions of the synthesized and GT images. We employ DINOv3~\cite{simeoni2025dinov3} as the feature extraction backbone for this purpose, as its dense and semantically discriminative representations enable fine‑grained structural correspondence.
Inspired by the multi-view consistency metric in previous work~\cite{elbanani2024probing}, we design a structure correspondence objective to match semantic patches between synthesized and GT images, ensuring local structure alignment, as shown in Fig.~\ref{fig:loss}(b). Specifically, both images are rescaled to a unified  resolution~$m \times m$ to obtain finer‑grained patches and processed by the DINOv3 backbone~$\psi(\cdot)$ to generate patch-level feature maps:
\begin{equation}
F_p = \psi(I_{\text{syn}}),\quad  F_q =\psi(I_{\text{GT}}),
\end{equation}
which are flattened into token sets $P=\{f_p\}$ and $Q=\{f_q\}$. 
For each predicted token~$f_p$, we find the most semantically matched target token~$f_q$ by maximizing their similarity. 
The semantic structure matching loss is formulated as:
\begin{equation}
\mathcal{L}_{\text{match}}
= 1 - \frac{1}{|P|}\sum_{p\in P}\max_{q\in Q}\langle f_p, f_q \rangle,
\label{eq:structure}
\end{equation}
where $\langle \cdot , \cdot \rangle$ denotes the cosine similarity between two feature tokens.
In this way, each region of the synthesized image $I_{\text{syn}}$ is matched with its most semantically similar counterpart in the GT image $I_{\text{GT}}$ within the semantic feature space, establishing fine-grained structural correspondence.

Finally, we can define our realism loss as:
\begin{equation}
\mathcal{L}_{\text{real}}
=
\mathcal{L}_{\text{adapt}}
+\mathcal{L}_{\text{match}},
\label{eq:total}
\end{equation}
which facilitates both realistic detail enhancement and semantic structure alignment.

\subsection{Paradigm‑Specific Training}
\label{sec:strategy}
While we introduce a general realistic detail‑enhancement scheme, its application to different 3D‑native generative models needs paradigm‑specific training designs.
Accordingly, as illustrated in the middle of part of Fig.~\ref{fig:pipeline}, we develop distinct training strategies for geometry–texture coupled and decoupled 3D‑native generation paradigms.

\vspace{+2mm}
\noindent\textbf{Coupled 3D-Native Generation}.
For geometry–texture coupled 3D-native generation, we choose the representative diffusion‑based generator Trellis~\cite{xiang2025structured} for training.
Trellis encodes 3D models into structured 3D latents and learns their distributions through a 3D diffusion process.
Since four views are insufficient to texture a 3D model, which is further analysed in the \textbf{Appendix~\ref{sec:analyse}}, we design a novel strategy to integrate realistic detail priors into the diffusion training without relying on GT 3D latents.
%As our Photo3D‑MV dataset is constructed with Trellis, we reuse its intermediate structured 3D latents, which remain structurally aligned with the realistic multi‑view images.

As illustrated in Fig.~\ref{fig:pipeline}(a), we first sample a structured 3D latent $x_0$ and perturb it with Gaussian noise to obtain a noised latent $x_t$.
We take the corresponding realistic multi‑views as GT images $I_{\text{GT}}$ and randomly select a single view as the condition image $I_{\text{cond}}$.
The noised latent $x_t$ and condition image $I_{\text{cond}}$ are fed into a 3D diffusion model parameterized by $\theta$ to predict the clean latent $\hat{x}_0$, which is subsequently decoded by a 3D decoder $D{\phi}$ into a 3DGS model. The 3DGS model is rendered at the same viewpoints as $I_{\text{GT}}$ and supervised with our realism loss $\mathcal{L}_{\text{real}}$, formulated as:

\begin{equation}
\begin{aligned}
\min_{\theta,\,\phi}\;
\mathbb{E}_{x_0,\,t\,I_{\text{cond}}}
\Big[
\mathcal{L}_{\text{real}}\big(
D_{\phi}(\hat{x}_0),\, I_{\text{GT}}
\big)
\Big], \\
\quad
\hat{x}_0 = x_t - t\,v_{\theta}(x_t,\,t,\,I_{\text{cond}}),
\end{aligned}
\label{eq:train_rectified}
\end{equation}
where $v_{\theta}$ represents the learned 
velocity field that rectifies the noisy 3D latent $x_t$ toward a clean latent $\hat{x}_0$. Although the original image condition of the $x_0$ differs from the current input $I_{\text{cond}}$, adding noise to $x_0$ encourages the diffusion model to explore an expanded generative space toward realism derived from the paired realistic GT images.

\begin{table*}[t]
\caption{\textbf{Comparison on the ImageNet and Real 3D datasets.}
Quantitative metrics evaluate input fidelity (CLIP, KID),
detail realism (MANIQA, MUSIQ),
and overall aesthetic quality (NIMA, Aesthetic Score).
Qualitative metrics include the Gemini‑2.5-based winning rate and human‑rated realism scores (1–5 scale, higher indicates greater realism) on 20 objects for each method.}
\label{tab:imagenet_realworld_results}
\centering
\small
\setlength{\tabcolsep}{4pt}
\renewcommand{\arraystretch}{1.1}
\begin{tabular}{@{}lccccccc|cc@{}}
\toprule
\multirow{3}{*}{\textbf{Method}} &
\multicolumn{7}{c|}{\textbf{Quantitative Metrics}} &
\multicolumn{2}{c}{\textbf{Qualitative Metrics}} \\
\cmidrule(lr){2-8} \cmidrule(lr){9-10}
& \multicolumn{2}{c}{\textbf{Fidelity}} 
& \multicolumn{2}{c}{\textbf{Realism}} 
& \multicolumn{3}{c|}{\textbf{Aesthetic Quality}} 
& \textbf{Gemini} & \textbf{Human} \\
\cmidrule(lr){2-3} \cmidrule(lr){4-5} \cmidrule(lr){6-8} \cmidrule(lr){9-9} \cmidrule(lr){10-10}
& CLIP$\uparrow$ & KID$\downarrow$ 
& MANIQA$\uparrow$ & MUSIQ$\uparrow$
& NIMA$\uparrow$ & Aes.$\uparrow$ & 
& Winning Rate\%$\uparrow$ & Score$\uparrow$ \\
\midrule
\multicolumn{10}{c}{\textbf{ImageNet~\cite{imagenet15russakovsky}}} \\
\midrule
Real3D & 0.600 & 0.074 & 0.261 & 29.666 & 4.185 & 3.843 &  & 0 & 1.0 \\
3DTopia-XL & 0.646 & 0.080 & 0.420 & 63.576 & 4.913 & 4.150 &  & 42.5 & 2.3 \\
Hunyuan3D & 0.645 & 0.058 & 0.458 & 69.396 & 5.182 & 4.558 &  & 55.0 & 2.5 \\

\midrule
Trellis & 0.672 & 0.045 & 0.438 & 69.108 & 5.239 & 4.682 &  & 68.1 & 3.4 \\
\textbf{Photo3D (Trellis)} & \textbf{\textcolor{red}{0.679}} & \textbf{\textcolor{red}{0.044}} & \textbf{\textcolor{red}{0.470}} & \textbf{\textcolor{red}{72.385}} & \textbf{\textcolor{red}{5.548}} & \textbf{\textcolor{red}{4.927}} &&  \textbf{\textcolor{red}{95.0}} & \textbf{\textcolor{red}{4.4}} \\
\midrule
Step1X‑3D & 0.672 & 0.046 & 0.438 & 70.471 & 5.387 & 4.701 &  & 66.9 & 3.4 \\
\textbf{Photo3D (Step1X-3D)} & \textbf{0.674} & \textbf{0.045} & \textbf{0.447} & \textbf{72.215} & \textbf{5.412} & \textbf{4.776} &  & \textbf{80.0} & \textbf{3.8} \\
\midrule
TexGaussian & 0.607 &0.104 &0.412 & 65.003 & 4.611 & 4.197 &  & 13.8 & 1.4 \\
\textbf{Photo3D (TexGaussian)} & \textbf{0.638} &  \textbf{0.082} & \textbf{0.445} & \textbf{70.273} & \textbf{5.151} & \textbf{4.576} &  & \textbf{28.8} &\textbf{ 3.0} \\
\midrule
\multicolumn{10}{c}{\textbf{Real 3D Datasets (GSO~\cite{downs2022google}, Omni3D~\cite{brazil2023omni3d}, DTC~\cite{dong2025digital})}} \\
\midrule
Real3D & 0.732 &  0.047 &  0.259 &30.004 & 4.066 & 3.684 &  & 0 & 1.0 \\
3DTopia-XL & 0.831 & 0.015 & 0.382 & 57.085 & 4.484 & 4.082 &  & 46.9 & 2.1 \\

Hunyuan3D &  0.811 & 0.006 &0.440 &  63.585 &4.607 & 4.524 &  & 50.6& 2.5 \\
\midrule
Trellis & 0.853 &  \textbf{0.002} &  0.427 & 64.155 & 4.653 & 4.481 &  & 70.6 & 3.9 \\
\textbf{Photo3D (Trellis)} & \textbf{\textcolor{red}{0.864} }& \textbf{\textcolor{red}{0.002}} &\textbf{\textcolor{red}{0.459}}& \textbf{\textcolor{red}{65.724}} & \textbf{\textcolor{red}{4.856}} & \textbf{\textcolor{red}{4.689}} &  & \textbf{\textcolor{red}{93.8}} & \textbf{\textcolor{red}{4.8}} \\
\midrule
Step1X‑3D & 0.849 & \textbf{0.004} & 0.416 & 63.132 & 4.713 & 4.516 &  & 65.0 & 3.1 \\
\textbf{Photo3D (Step1X‑3D)} & \textbf{0.856} & \textbf{0.004} &  \textbf{0.424} &\textbf{64.904} &\textbf{4.738} &  \textbf{4.593} &  & \textbf{83.1} & \textbf{3.6} \\
\midrule
TexGaussian &  0.749 & \textbf{0.039} & 0.441 &61.837 & 4.440 &4.346 &  & 14.4 & 1.7 \\
\textbf{Photo3D (TexGaussian)} & \textbf{0.766} & \textbf{0.039} & \textbf{0.444} & \textbf{64.706} & \textbf{4.642} & \textbf{4.531} &  & \textbf{25.6} & \textbf{1.8} \\
\bottomrule
\end{tabular}
\vspace{-2mm}
\end{table*}

\vspace{+2mm}
\noindent\textbf{Decoupled 3D-Native Generation}.
For the geometry–texture decoupled paradigm, we design distinct training strategies for 3D‑native and multi‑view texturing models based on 3D‑native geometry.
We first employ TexGaussian~\cite{xiong2025texgaussian}, a single feed‑forward 3D-native texturing model that generates an octree‑based 3DGS directly from a given 3D mesh in one pass, serving as the GT for texture optimization of the 3D mesh.
As shown in Fig.~\ref{fig:pipeline}(b), we train the texturing model $T_{\theta}$, conditioned on a 3D mesh $\mathcal{M}$ from Photo3D‑MV and a text description $y_{\text{text}}$, to generate a textured 3DGS, which is then rendered at the same viewpoints as $I_{\text{GT}}$ and supervised by $\mathcal{L}_{\text{real}}$, defined as:
\begin{equation}
\begin{aligned}
\min_{\theta}\;
\mathbb{E}_{\mathcal{M},\,y_{\text{text}}}
\Big[
\mathcal{L}_{\text{real}}\big(
T_{\theta}(y_{\text{text}},\mathcal{M}),\, I_{\text{GT}}
\big)
\Big].
\end{aligned}
\label{eq:train_texgaussian}
\end{equation}
This single feed‑forward 3D‑native texturing process enables efficient and direct synthesis of photorealistic and geometrically consistent appearances on 3D‑native geometry.

%Within the geometry–texture decoupled paradigm, another category of texturing models follows a diffusion‑based 
For multi‑view generation methods, we adopt the representative Step1X-3D texturing model~\cite{li2025step1x}, which is built upon the pretrained single‑image diffusion model~\cite{lin2024common} and fine‑tuned on 3D datasets for multi‑view generation.
However, the fine‑tuned model often produces over-smoothed textures, thereby reducing the realism and fine details in the generated textures. To solve this problem, as shown in Fig.~\ref{fig:pipeline}(c), we encode the realistic multi‑view images from Photo3D‑MV into multi‑view latents $X_0 = \{x_i\}_{i=1}^4$, where each $x_i$ corresponds to a single‑view image, and add Gaussian noise to $X_0$ to obtain the noised latents $X_t$.  
The diffusion‑based model takes $X_t$ as input and the rendered geometry images of the 3D mesh $\mathcal{M}$ as the condition, predicts the noise $\epsilon_{\theta}$, and restores the clean latents $\hat{X}_0$. The frozen 2D decoder $D_{\phi}$ from the texturing model then decodes $\hat{X}_0$ into images for our realism supervision as:
\begin{equation}
\begin{aligned}
\min_{\theta}\;
\mathbb{E}_{X_0,\,t,\,I_{\text{cond}}}
\Big[
\mathcal{L}_{\text{real}}\big(
D_{\phi}(\hat{X}_0),\, I_{\text{GT}}
\big)
\Big], \\ \quad
\hat{X}_0
= \alpha_t\, X_t - \beta_t\, \epsilon_{\theta}(X_t,\,t,\,I_{\text{cond}},\mathcal{M}),
\end{aligned}
\label{eq:train_mv_ddpm}
\end{equation}
where $\alpha_t$ and $\beta_t$ are the noise scheduling coefficients. With this training strategy, we are able to advance the multi‑view texturing models with high-fidelity realistic details.

\section{Experiments}
\subsection{Experiment Setup}
\textbf{Baseline Models and Fine-tuning Details}.
In this paper, we adopt GPT-4o-Image~\cite{hurst2024gpt} as the 2D generator for constructing the Photo3D-MV dataset. However, our framework is not restricted to this specific model and can be readily extended to other 2D generative models, as shown in \textbf{Appendix~\ref{sec:choice}}.

We use the proposed Photo3D‑MV dataset to finetune three representative 3D generators as our Photo3D models.
The first is Trellis \cite{xiang2025structured}, a geometry–texture coupled 3D‑native generator, while the other two are geometry–texture decoupled approaches, Step1X‑3D~\cite{li2025step1x}, and TexGaussian \cite{xiong2025texgaussian}. 
%With these baselines, we further develop three Photo3D generation models built upon Trellis~\cite{xiang2025structured}, Step1X-3D~\cite{li2025step1x}, and TexGaussian~\cite{xiong2025texgaussian}, respectively. 
Since TexGaussian is a texturing model, we couple it with the 3D geometries generated by Step1X‑3D. 

We fine-tune these models on 8 NVIDIA H20 GPUs following their original training and inference hyperparameters with the AdamW optimizer~\cite{loshchilov2017decoupled}. Specifically, we fine‑tune Trellis~\cite{xiang2025structured} (1.1B parameters) for 100K steps with a batch size of 1 and a learning rate of $1\times10^{-4}$.
For TexGaussian~\cite{xiong2025texgaussian}, we fine‑tune its Objaverse‑trained text‑conditioned model for 10K steps with a batch size of 5 and a learning rate of $4\times10^{-4}$.
For Step1X‑3D~\cite{li2025step1x}, we fine‑tune its texturing model for 10K steps with a batch size of 1 and a learning rate of $1\times10^{-4}$. 
We set both the input and output image resolutions to $512\times512$. For computing $L_{\text{match}}$, we set the upsampling resolution to $m=1024$. 

\vspace{+1mm}
\noindent\textbf{Compared Methods and Evaluation Datasets}.
We compare our fine-tuned 3D generators with their original counterparts, as well as several advanced models, including Real3D~\cite{Jiang2025ICCV}, an approach emphasizing realistic 3D generation based on the large 3D reconstruction model~\cite{hong2023lrm}, 3DTopia‑XL~\cite{chen20253dtopia} and Hunyuan3D 2.0~\cite{zhao2025hunyuan3d}.

For evaluation, we employ two types of test sets.
First, we select 400 high‑quality samples based on  aesthetic scores~\cite{improved_aesthetic_predictor} from real 3D scanned datasets, including Google Scanned Objects (GSO)~\cite{downs2022google}, OmniObject3D (Omni3D)~\cite{brazil2023omni3d}, and Digital Twin Catalog (DTC)~\cite{dong2025digital}, as the test set.
Considering that these scanned 3D data have limited category diversity and often exhibit insufficient fine‑details, we further select 1,000 images with top aesthetic scores from the ImageNet~\cite{imagenet15russakovsky} dataset as another test set.
Since ImageNet images often contain occlusions, we use Gemini‑2.5‑Flash~\cite{comanici2025gemini} to extract clean and complete foregrounds.
Besides, we use BLIP‑2~\cite{li2023blip} to generate paired descriptions as text input prompts.

\begin{figure*}[t]
  \centering
  \includegraphics[width=1\linewidth]{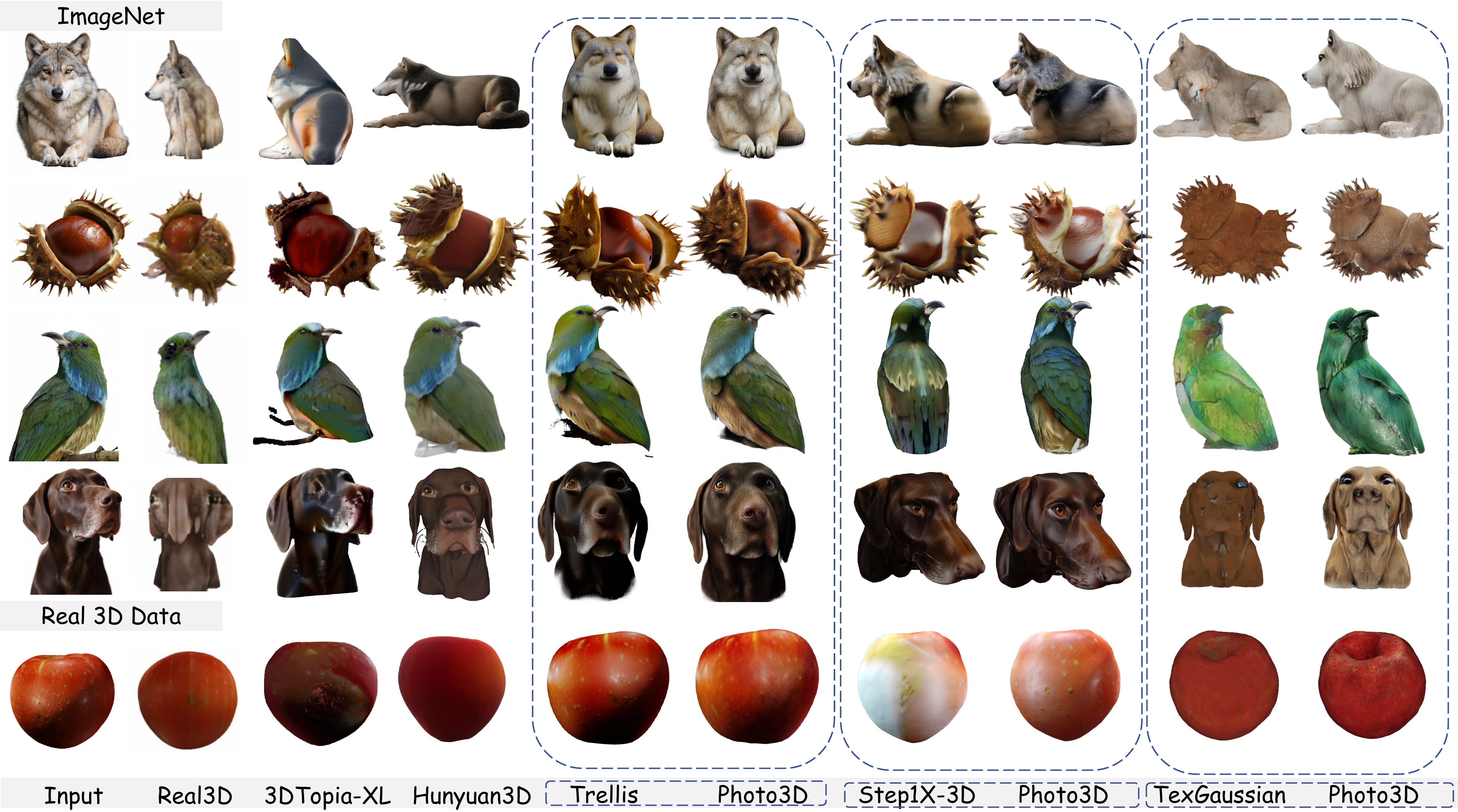}
  \vspace{-6mm}
 \caption{Visual comparison of generated 3D assets between existing methods and our Photo3D models built upon different 3D generation paradigms.
For TexGaussian and Photo3D (TexGaussian), we use the generated captions of the input images as text conditions.}
  \label{fig:duibi}
  \vspace{-3mm}
\end{figure*}

\vspace{+1mm}
\noindent\textbf{Evaluation Metrics}.
We quantitatively evaluate the fidelity, realism, and aesthetic quality of the generated 3D assets.
Fidelity is evaluated using CLIP similarity~\cite{radford2021learning} and Kernel Inception Distance (KID)~\cite{binkowski2018demystifying}, measuring the alignment between multi‑view renderings and their input images.
Realism is measured with MANIQA~\cite{yang2022maniqa} and MUSIQ~\cite{ke2021musiq}, which estimate the realism of rendered images.
The overall aesthetic quality is assessed using NIMA~\cite{talebi2018nima} and the aesthetic score~\cite{improved_aesthetic_predictor}.
We further qualitatively evaluate the 3D perceptual realism using Gemini‑2.5~\cite{comanici2025gemini} and human evaluation.
Gemini‑2.5 performs pairwise comparisons to compute the winning rate among different 3D models, while the user study involves 20 participants to rate each model on a 1–5 scale, where higher scores indicate greater realism.

\subsection{Experimental Results}
We compare Photo3D with the competing methods in Tab. \ref{tab:imagenet_realworld_results} and Fig. \ref{fig:duibi}. Two observations can be made. First, across different 3D generation paradigms, Photo3D consistently improves its baseline model in both quantitative and qualitative metrics. Second, the Photo3D model built upon Trellis achieves the best performance among all models.

In particular, as shown in Tab.~\ref{tab:imagenet_realworld_results}, Photo3D effectively enhances the photorealistic performance, leading to more faithful generation for real‑image inputs and correspondingly higher fidelity scores. Specifically, it better retains fine 3D structural details and maintains cross‑view perceptual coherence, yielding realistic views aligned with the appearance of the conditioned input perspectives.
For realism assessment, the consistently higher realism scores validate Photo3D's capacity to enrich realistic fine‑grained details.
In terms of visual aesthetic quality, Photo3D‑MV offers diverse and photorealistic multi‑view supervision with high aesthetic scores. Leveraging this prior, Photo3D produces finer material details, more natural shading, and structure-aligned textures, leading to visually superior 3D model generation. Qualitatively, both Gemini and human evaluations confirm that Photo3D delivers more realistic 3D generation.

\begin{table}[t]
\caption{\textbf{Ablation study on different supervision types.} We analyze the effect of each component of our method and other supervision approaches on Photo3D (Trellis).}
\label{tab:ablation_losses}
\vspace{-2mm}
\centering
\small
\resizebox{\columnwidth}{!}{%
\setlength{\tabcolsep}{4pt}
\renewcommand{\arraystretch}{1.1}
\begin{tabular}{@{}lccccccc@{}}
\toprule
\multirow{2}{*}{\textbf{Method}} &
\multicolumn{2}{c}{\textbf{Fidelity}} &
\multicolumn{2}{c}{\textbf{Realism}} &
\multicolumn{2}{c}{\textbf{Aesthetic Quality}} \\
\cmidrule(lr){2-3}\cmidrule(lr){4-5}\cmidrule(lr){6-7}
& CLIP$\uparrow$ & KID$\downarrow$ &
MANIQA$\uparrow$ & MUSIQ$\uparrow$ &
NIMA$\uparrow$ & Aes.$\uparrow$ \\
\midrule
\textbf{Ours} & \textbf{0.679} & \textbf{0.044} & \textbf{0.470} & \textbf{72.385} & \textbf{5.548} & \textbf{4.927}  \\
w/o $L_{\text{adapt}}$ &0.668 & 0.054&0.308 & 61.681& 4.726&4.579 \\
w/o $L_{\text{match}}$ & 0.671& 0.048& 0.409& 71.069& 5.400&4.897 \\
w/o all (Baseline) &0.672 & 0.045& 0.438&69.108 &5.239 &4.682 \\
\midrule
\multicolumn{7}{c}{\textbf{With Other Realistic Enhancing Supervision Types}} \\  
\midrule
w/ $L_2$ loss &0.598 &0.195 &0.346 & 55.281&4.782 & 4.225\\
w/ Gram loss &0.671 & 0.049&0.444 &69.646 &5.241 &4.682 \\
w/ GAN loss &0.672 &0.046 &0.468
 &72.058 &5.260 &4.901 \\
\bottomrule
\end{tabular}
}% end resizebox
\vspace{-2mm}
\end{table}

We further visualize the generated 3D results in Fig.~\ref{fig:duibi}. We can see that the Real3D model generally fails to synthesize plausible novel views, whereas 3DTopia‑XL and Hunyuan3D often produce smooth and synthetic textures, possibly due to the domain bias introduced by their reliance on predominantly synthetic 3D training datasets. More detailed visualizations are in the \textbf{Appendix~\ref{sec:morere}}. 

\vspace{+1mm}
\noindent\textbf{Analyses}.
We then analyze the 3D generation pipelines of our fine-tuned models, including Trellis, Step1X‑3D and TexGaussian, together with their corresponding original counterparts. First, for the geometry–texture coupled pipeline represented by Trellis, it produces 3D consistent textures on 3D-native geometry but still exhibits unrealistic appearances. For example, in the wolf example, the fur texture lacks fine‑grained hair strands and manifests as coarse granular patterns, which can be attributed to the absence of realistic exemplars in the training 3D datasets. In contrast, Photo3D (Trellis) generates highly detailed and perceptually realistic textures, achieving stronger fidelity to the input images.
Step1X‑3D adopts the geometry–texture decoupled paradigm, where the 3D geometry is first generated and then textured using a multi‑view generation model. Since it is trained on synthetic-dominant 3D datasets, the textures often exhibit distortions. Photo3D (Step1X‑3D) alleviates these issues by being further trained on realistic multi‑view images, producing more faithful results, such as the chestnut example. 
TexGaussian is a 3D-native texturing model, and we apply it to the 3D geometries generated by Step1X‑3D to evaluate its performance. Although it generates 3D‑consistent textures through a single feed‑forward pass, its limited generative capacity results in a lack of fine‑grained details. Photo3D (TexGaussian) addresses these limitations by enriching the textures with more natural surface variations and refined details, producing perceptually more realistic 3D results.

Overall, Photo3D consistently delivers more realistic 3D generations across diverse 3D-native generation paradigms.

\subsection{Ablation Studies}

\begin{figure}[t]
  \centering
  \includegraphics[width=1\linewidth]{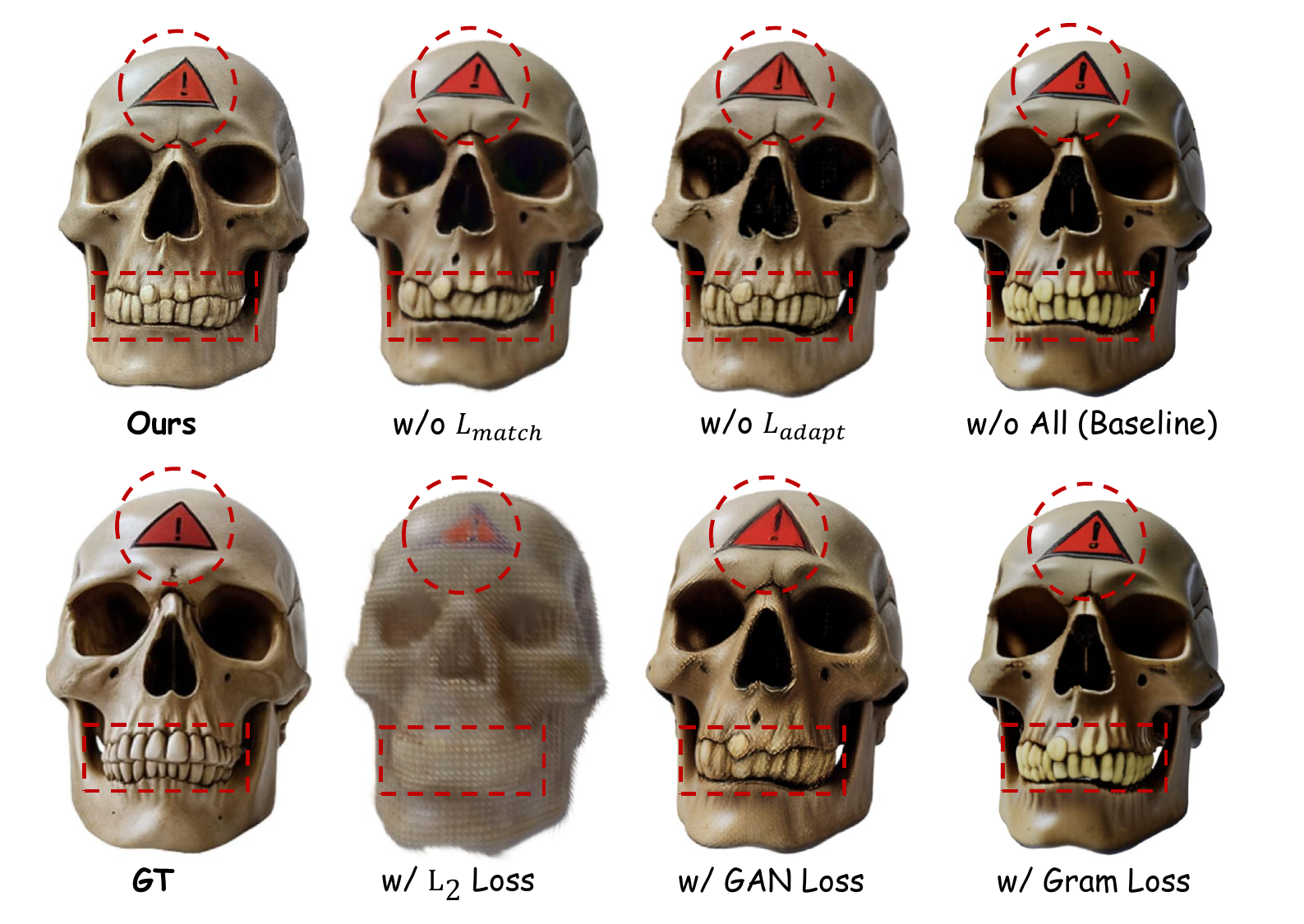}
  \vspace{-6mm}
\caption{Ablation results on rendered images of generated 3D models. Highlighted regions mark areas with noticeable detail and color variations for comparison.}
  \label{fig:ablation}
  \vspace{-3mm}
\end{figure}

We then perform an ablation study on the supervision types of Photo3D using the Trellis‑based model. Specifically, we compare the effects of different supervision losses on the generated 3D quality. The results are presented in Tab.~\ref{tab:ablation_losses} and Fig.~\ref{fig:ablation}.
We see that removing $L_{\text{adapt}}$, which performs perceptual feature adaptation, and keeping only $L_{\text{match}}$ for semantic structural matching can preserve stable 3D structure but result in low‑resolution textures caused by patch‑level matching.
Conversely, removing $L_{\text{match}}$ while retaining $L_{\text{adapt}}$ leads to texture misalignment and noticeable structural drift in the generated 3D models.

In addition, we compare other supervision approaches for realism enhancement.
A simple $L_2$ loss causes severe texture distortion and even collapse of the generated 3D model.
The GAN loss~\cite{goodfellow2014generative} transfers certain visual realism priors but fails to maintain semantic structure consistency.
The style‑transfer Gram loss~\cite{gatys2015texture} focuses on global texture patterns rather than spatial or lighting realism, leading to limited improvement in 3D realism.
These results demonstrate the effectiveness and well‑founded design of our realistic detail enhancement scheme.

\section{Conclusion}
In this work, we presented Photo3D, a framework for photorealistic 3D generation through structure‑aligned detail enhancement. We first introduced a 3D‑aligned multi‑view synthesis pipeline and built a realistic multi‑view dataset, Photo3D‑MV. We then proposed a general detail enhancement scheme with specific training strategies for different 3D‑native generation paradigms. Experiments showed that Photo3D could consistently and significantly improve the photorealistic appearance of the 3D models generated by its original counterpart, achieving state‑of‑the‑art photorealistic 3D generation performance.
This highlighted the potential of leveraging 2D knowledge to complement costly and limited 3D assets, enabling advanced 3D generation.

\vspace{+1mm}
\noindent\textbf{Limitations}.
While Photo3D produces photorealistic 3D models using realistic detail priors from GPT‑4o‑Image, the generated 3D appearance may inherit its generation preferences. This bias can be reduced with the evolving advances in image generators, which will be our future work.

{
    \small
    \bibliographystyle{ieeenat_fullname}
    \bibliography{main}

\begin{thebibliography}{77}
\providecommand{\natexlab}[1]{#1}
\providecommand{\url}[1]{\texttt{#1}}
\expandafter\ifx\csname urlstyle\endcsname\relax
  \providecommand{\doi}[1]{doi: #1}\else
  \providecommand{\doi}{doi: \begingroup \urlstyle{rm}\Url}\fi

\bibitem[gem(2025)]{gemini3pro}
Nano banana pro.
\newblock \url{https://deepmind.google/models/gemini/pro/}, 2025.

\bibitem[AI@Meta(2024)]{llama3modelcard}
AI@Meta.
\newblock Llama 3 model card.
\newblock 2024.

\bibitem[Bensadoun et~al.(2024)Bensadoun, Kleiman, Azuri, Harosh, Vedaldi, Neverova, and Gafni]{bensadoun2024meta}
Raphael Bensadoun, Yanir Kleiman, Idan Azuri, Omri Harosh, Andrea Vedaldi, Natalia Neverova, and Oran Gafni.
\newblock Meta 3d texturegen: Fast and consistent texture generation for 3d objects.
\newblock \emph{arXiv preprint arXiv:2407.02430}, 2024.

\bibitem[Bi{\'n}kowski et~al.(2018)Bi{\'n}kowski, Sutherland, Arbel, and Gretton]{binkowski2018demystifying}
Miko{\l}aj Bi{\'n}kowski, Danica~J Sutherland, Michael Arbel, and Arthur Gretton.
\newblock Demystifying mmd gans.
\newblock \emph{arXiv preprint arXiv:1801.01401}, 2018.

\bibitem[Brazil et~al.(2023)Brazil, Kumar, Straub, Ravi, Johnson, and Gkioxari]{brazil2023omni3d}
Garrick Brazil, Abhinav Kumar, Julian Straub, Nikhila Ravi, Justin Johnson, and Georgia Gkioxari.
\newblock Omni3d: A large benchmark and model for 3d object detection in the wild.
\newblock In \emph{Proceedings of the IEEE/CVF conference on computer vision and pattern recognition}, pages 13154--13164, 2023.

\bibitem[Chang et~al.(2015)Chang, Funkhouser, Guibas, Hanrahan, Huang, Li, Savarese, Savva, Song, Su, et~al.]{chang2015shapenet}
Angel~X Chang, Thomas Funkhouser, Leonidas Guibas, Pat Hanrahan, Qixing Huang, Zimo Li, Silvio Savarese, Manolis Savva, Shuran Song, Hao Su, et~al.
\newblock Shapenet: An information-rich 3d model repository.
\newblock \emph{arXiv preprint arXiv:1512.03012}, 2015.

\bibitem[Chen et~al.(2024{\natexlab{a}})Chen, Xu, Esposito, Tang, and Geiger]{chen2024lara}
Anpei Chen, Haofei Xu, Stefano Esposito, Siyu Tang, and Andreas Geiger.
\newblock Lara: Efficient large-baseline radiance fields.
\newblock In \emph{European Conference on Computer Vision}, pages 338--355. Springer, 2024{\natexlab{a}}.

\bibitem[Chen et~al.(2025{\natexlab{a}})Chen, Li, Zhang, Wang, and Zhang]{chen2025fast}
Liyi Chen, Ruihuang Li, Guowen Zhang, Pengfei Wang, and Lei Zhang.
\newblock Fast multi-view consistent 3d editing with video priors.
\newblock \emph{arXiv preprint arXiv:2511.23172}, 2025{\natexlab{a}}.

\bibitem[Chen et~al.(2023)Chen, Chen, Jiao, and Jia]{chen2023fantasia3d}
Rui Chen, Yongwei Chen, Ningxin Jiao, and Kui Jia.
\newblock Fantasia3d: Disentangling geometry and appearance for high-quality text-to-3d content creation.
\newblock In \emph{Proceedings of the IEEE/CVF international conference on computer vision}, pages 22246--22256, 2023.

\bibitem[Chen et~al.(2024{\natexlab{b}})Chen, Wang, Wang, and Liu]{chen2024text}
Zilong Chen, Feng Wang, Yikai Wang, and Huaping Liu.
\newblock Text-to-3d using gaussian splatting.
\newblock In \emph{Proceedings of the IEEE/CVF conference on computer vision and pattern recognition}, pages 21401--21412, 2024{\natexlab{b}}.

\bibitem[Chen et~al.(2025{\natexlab{b}})Chen, Tang, Dong, Cao, Hong, Lan, Wang, Xie, Wu, Saito, et~al.]{chen20253dtopia}
Zhaoxi Chen, Jiaxiang Tang, Yuhao Dong, Ziang Cao, Fangzhou Hong, Yushi Lan, Tengfei Wang, Haozhe Xie, Tong Wu, Shunsuke Saito, et~al.
\newblock 3dtopia-xl: Scaling high-quality 3d asset generation via primitive diffusion.
\newblock In \emph{Proceedings of the Computer Vision and Pattern Recognition Conference}, pages 26576--26586, 2025{\natexlab{b}}.

\bibitem[Collins et~al.(2022)Collins, Goel, Deng, Luthra, Xu, Gundogdu, Zhang, Vicente, Dideriksen, Arora, et~al.]{collins2022abo}
Jasmine Collins, Shubham Goel, Kenan Deng, Achleshwar Luthra, Leon Xu, Erhan Gundogdu, Xi Zhang, Tomas F~Yago Vicente, Thomas Dideriksen, Himanshu Arora, et~al.
\newblock Abo: Dataset and benchmarks for real-world 3d object understanding.
\newblock In \emph{Proceedings of the IEEE/CVF conference on computer vision and pattern recognition}, pages 21126--21136, 2022.

\bibitem[Comanici et~al.(2025)Comanici, Bieber, Schaekermann, Pasupat, Sachdeva, Dhillon, Blistein, Ram, Zhang, Rosen, et~al.]{comanici2025gemini}
Gheorghe Comanici, Eric Bieber, Mike Schaekermann, Ice Pasupat, Noveen Sachdeva, Inderjit Dhillon, Marcel Blistein, Ori Ram, Dan Zhang, Evan Rosen, et~al.
\newblock Gemini 2.5: Pushing the frontier with advanced reasoning, multimodality, long context, and next generation agentic capabilities.
\newblock \emph{arXiv preprint arXiv:2507.06261}, 2025.

\bibitem[Deitke et~al.(2023{\natexlab{a}})Deitke, Liu, Wallingford, Ngo, Michel, Kusupati, Fan, Laforte, Voleti, Gadre, et~al.]{deitke2023objaversexl}
Matt Deitke, Ruoshi Liu, Matthew Wallingford, Huong Ngo, Oscar Michel, Aditya Kusupati, Alan Fan, Christian Laforte, Vikram Voleti, Samir~Yitzhak Gadre, et~al.
\newblock Objaverse-xl: A universe of 10m+ 3d objects.
\newblock \emph{Advances in Neural Information Processing Systems}, 36:\penalty0 35799--35813, 2023{\natexlab{a}}.

\bibitem[Deitke et~al.(2023{\natexlab{b}})Deitke, Schwenk, Salvador, Weihs, Michel, VanderBilt, Schmidt, Ehsani, Kembhavi, and Farhadi]{deitke2023objaverse}
Matt Deitke, Dustin Schwenk, Jordi Salvador, Luca Weihs, Oscar Michel, Eli VanderBilt, Ludwig Schmidt, Kiana Ehsani, Aniruddha Kembhavi, and Ali Farhadi.
\newblock Objaverse: A universe of annotated 3d objects.
\newblock In \emph{Proceedings of the IEEE/CVF conference on computer vision and pattern recognition}, pages 13142--13153, 2023{\natexlab{b}}.

\bibitem[DmenuACl and HS()]{dmenuaclcommission}
Ernst DmenuACl and R HS.
\newblock Commission internationale de l'eclairage c.

\bibitem[Dong et~al.(2025)Dong, Chen, Lv, Yu, Zhang, Zhang, Zhu, Tian, Li, Moffatt, et~al.]{dong2025digital}
Zhao Dong, Ka Chen, Zhaoyang Lv, Hong-Xing Yu, Yunzhi Zhang, Cheng Zhang, Yufeng Zhu, Stephen Tian, Zhengqin Li, Geordie Moffatt, et~al.
\newblock Digital twin catalog: A large-scale photorealistic 3d object digital twin dataset.
\newblock In \emph{Proceedings of the Computer Vision and Pattern Recognition Conference}, pages 753--763, 2025.

\bibitem[Downs et~al.(2022)Downs, Francis, Koenig, Kinman, Hickman, Reymann, McHugh, and Vanhoucke]{downs2022google}
Laura Downs, Anthony Francis, Nate Koenig, Brandon Kinman, Ryan Hickman, Krista Reymann, Thomas~B McHugh, and Vincent Vanhoucke.
\newblock Google scanned objects: A high-quality dataset of 3d scanned household items.
\newblock In \emph{2022 International Conference on Robotics and Automation (ICRA)}, pages 2553--2560. IEEE, 2022.

\bibitem[El~Banani et~al.(2024)El~Banani, Raj, Maninis, Kar, Li, Rubinstein, Sun, Guibas, Johnson, and Jampani]{elbanani2024probing}
Mohamed El~Banani, Amit Raj, Kevis-Kokitsi Maninis, Abhishek Kar, Yuanzhen Li, Michael Rubinstein, Deqing Sun, Leonidas Guibas, Justin Johnson, and Varun Jampani.
\newblock {Probing the 3D Awareness of Visual Foundation Models}.
\newblock In \emph{CVPR}, 2024.

\bibitem[Fang et~al.(2024)Fang, Sun, Wu, Wang, Liu, Wetzstein, and Lin]{fang2024make}
Ye Fang, Zeyi Sun, Tong Wu, Jiaqi Wang, Ziwei Liu, Gordon Wetzstein, and Dahua Lin.
\newblock Make-it-real: Unleashing large multimodal model for painting 3d objects with realistic materials.
\newblock \emph{Advances in Neural Information Processing Systems}, 37:\penalty0 99262--99298, 2024.

\bibitem[Fei et~al.(2025)Fei, Tang, Tian, Shi, and Tan]{fei2025pacture}
Fan Fei, Jiajun Tang, Fei-Peng Tian, Boxin Shi, and Ping Tan.
\newblock Pacture: Efficient pbr texture generation on packed views with visual autoregressive models.
\newblock \emph{arXiv preprint arXiv:2505.22394}, 2025.

\bibitem[Gatys et~al.(2015)Gatys, Ecker, and Bethge]{gatys2015texture}
Leon Gatys, Alexander~S Ecker, and Matthias Bethge.
\newblock Texture synthesis using convolutional neural networks.
\newblock \emph{Advances in neural information processing systems}, 28, 2015.

\bibitem[Goodfellow et~al.(2014)Goodfellow, Pouget-Abadie, Mirza, Xu, Warde-Farley, Ozair, Courville, and Bengio]{goodfellow2014generative}
Ian~J Goodfellow, Jean Pouget-Abadie, Mehdi Mirza, Bing Xu, David Warde-Farley, Sherjil Ozair, Aaron Courville, and Yoshua Bengio.
\newblock Generative adversarial nets.
\newblock \emph{Advances in neural information processing systems}, 27, 2014.

\bibitem[Guo et~al.(2025)Guo, Gao, Bian, Sun, Zheng, Jia, and Gong]{guo2025hyper3d}
Jingyu Guo, Sensen Gao, Jia-Wang Bian, Wanhu Sun, Heliang Zheng, Rongfei Jia, and Mingming Gong.
\newblock Hyper3d: Efficient 3d representation via hybrid triplane and octree feature for enhanced 3d shape variational auto-encoders.
\newblock \emph{arXiv preprint arXiv:2503.10403}, 2025.

\bibitem[Gupta et~al.(2019)Gupta, Dollar, and Girshick]{gupta2019lvis}
Agrim Gupta, Piotr Dollar, and Ross Girshick.
\newblock Lvis: A dataset for large vocabulary instance segmentation.
\newblock In \emph{Proceedings of the IEEE/CVF conference on computer vision and pattern recognition}, pages 5356--5364, 2019.

\bibitem[Han et~al.(2024)Han, Wu, Shi, Liu, Liao, Qiu, Yuan, Gu, Dong, and Cui]{han2024mvimgnet2}
Xiaoguang Han, Yushuang Wu, Luyue Shi, Haolin Liu, Hongjie Liao, Lingteng Qiu, Weihao Yuan, Xiaodong Gu, Zilong Dong, and Shuguang Cui.
\newblock Mvimgnet2. 0: A larger-scale dataset of multi-view images.
\newblock \emph{arXiv preprint arXiv:2412.01430}, 2024.

\bibitem[He et~al.(2025)He, Yang, Yang, Tang, Wang, Zhang, Chen, Liu, Jiang, Guo, et~al.]{he2025materialmvp}
Zebin He, Mingxin Yang, Shuhui Yang, Yixuan Tang, Tao Wang, Kaihao Zhang, Guanying Chen, Yuhong Liu, Jie Jiang, Chunchao Guo, et~al.
\newblock Materialmvp: Illumination-invariant material generation via multi-view pbr diffusion.
\newblock \emph{arXiv preprint arXiv:2503.10289}, 2025.

\bibitem[Hong et~al.(2023)Hong, Zhang, Gu, Bi, Zhou, Liu, Liu, Sunkavalli, Bui, and Tan]{hong2023lrm}
Yicong Hong, Kai Zhang, Jiuxiang Gu, Sai Bi, Yang Zhou, Difan Liu, Feng Liu, Kalyan Sunkavalli, Trung Bui, and Hao Tan.
\newblock Lrm: Large reconstruction model for single image to 3d.
\newblock \emph{arXiv preprint arXiv:2311.04400}, 2023.

\bibitem[Huang et~al.(2025{\natexlab{a}})Huang, Wang, Liu, and Wang]{huang2025material}
Xin Huang, Tengfei Wang, Ziwei Liu, and Qing Wang.
\newblock Material anything: Generating materials for any 3d object via diffusion.
\newblock In \emph{Proceedings of the Computer Vision and Pattern Recognition Conference}, pages 26556--26565, 2025{\natexlab{a}}.

\bibitem[Huang et~al.(2025{\natexlab{b}})Huang, Guo, Wang, Yi, Ma, Cao, and Sheng]{huang2025mv}
Zehuan Huang, Yuan-Chen Guo, Haoran Wang, Ran Yi, Lizhuang Ma, Yan-Pei Cao, and Lu Sheng.
\newblock Mv-adapter: Multi-view consistent image generation made easy.
\newblock In \emph{Proceedings of the IEEE/CVF International Conference on Computer Vision}, pages 16377--16387, 2025{\natexlab{b}}.

\bibitem[Hurst et~al.(2024)Hurst, Lerer, Goucher, Perelman, Ramesh, Clark, Ostrow, Welihinda, Hayes, Radford, et~al.]{hurst2024gpt}
Aaron Hurst, Adam Lerer, Adam~P Goucher, Adam Perelman, Aditya Ramesh, Aidan Clark, AJ Ostrow, Akila Welihinda, Alan Hayes, Alec Radford, et~al.
\newblock Gpt-4o system card.
\newblock \emph{arXiv preprint arXiv:2410.21276}, 2024.

\bibitem[Jiang et~al.(2025)Jiang, Huang, and Pavlakos]{Jiang2025ICCV}
Hanwen Jiang, Qixing Huang, and Georgios Pavlakos.
\newblock Real3d: Towards scaling large reconstruction models with real images.
\newblock In \emph{Proceedings of the IEEE/CVF International Conference on Computer Vision (ICCV)}, pages 5821--5833, 2025.

\bibitem[Ke et~al.(2021)Ke, Wang, Wang, Milanfar, and Yang]{ke2021musiq}
Junjie Ke, Qifei Wang, Yilin Wang, Peyman Milanfar, and Feng Yang.
\newblock Musiq: Multi-scale image quality transformer.
\newblock In \emph{Proceedings of the IEEE/CVF international conference on computer vision}, pages 5148--5157, 2021.

\bibitem[Kerbl et~al.(2023)Kerbl, Kopanas, Leimk{\"u}hler, and Drettakis]{kerbl20233d}
Bernhard Kerbl, Georgios Kopanas, Thomas Leimk{\"u}hler, and George Drettakis.
\newblock 3d gaussian splatting for real-time radiance field rendering.
\newblock \emph{ACM Trans. Graph.}, 42\penalty0 (4):\penalty0 139--1, 2023.

\bibitem[Labs(2024)]{flux2024}
Black~Forest Labs.
\newblock Flux.
\newblock \url{https://github.com/black-forest-labs/flux}, 2024.

\bibitem[Labs(2025)]{flux-2-2025}
Black~Forest Labs.
\newblock {FLUX.2: Frontier Visual Intelligence}.
\newblock \url{https://bfl.ai/blog/flux-2}, 2025.

\bibitem[Labs et~al.(2025)Labs, Batifol, Blattmann, Boesel, Consul, Diagne, Dockhorn, English, English, Esser, Kulal, Lacey, Levi, Li, Lorenz, Müller, Podell, Rombach, Saini, Sauer, and Smith]{labs2025flux1kontextflowmatching}
Black~Forest Labs, Stephen Batifol, Andreas Blattmann, Frederic Boesel, Saksham Consul, Cyril Diagne, Tim Dockhorn, Jack English, Zion English, Patrick Esser, Sumith Kulal, Kyle Lacey, Yam Levi, Cheng Li, Dominik Lorenz, Jonas Müller, Dustin Podell, Robin Rombach, Harry Saini, Axel Sauer, and Luke Smith.
\newblock Flux.1 kontext: Flow matching for in-context image generation and editing in latent space, 2025.

\bibitem[Li et~al.(2023)Li, Li, Savarese, and Hoi]{li2023blip}
Junnan Li, Dongxu Li, Silvio Savarese, and Steven Hoi.
\newblock Blip-2: Bootstrapping language-image pre-training with frozen image encoders and large language models.
\newblock In \emph{International conference on machine learning}, pages 19730--19742. PMLR, 2023.

\bibitem[Li et~al.(2024)Li, Liu, Long, Zhang, Lin, Li, Qi, Zhang, Xue, Luo, et~al.]{li2024era3d}
Peng Li, Yuan Liu, Xiaoxiao Long, Feihu Zhang, Cheng Lin, Mengfei Li, Xingqun Qi, Shanghang Zhang, Wei Xue, Wenhan Luo, et~al.
\newblock Era3d: High-resolution multiview diffusion using efficient row-wise attention.
\newblock \emph{Advances in Neural Information Processing Systems}, 37:\penalty0 55975--56000, 2024.

\bibitem[Li et~al.(2025{\natexlab{a}})Li, Zhang, Sun, Qi, Li, Cheng, Cai, Wu, Liu, Wang, et~al.]{li2025step1x}
Weiyu Li, Xuanyang Zhang, Zheng Sun, Di Qi, Hao Li, Wei Cheng, Weiwei Cai, Shihao Wu, Jiarui Liu, Zihao Wang, et~al.
\newblock Step1x-3d: Towards high-fidelity and controllable generation of textured 3d assets.
\newblock \emph{arXiv preprint arXiv:2505.07747}, 2025{\natexlab{a}}.

\bibitem[Li et~al.(2025{\natexlab{b}})Li, Zou, Liu, Wang, Liang, Yu, Liu, Guo, Liang, Ouyang, et~al.]{li2025triposg}
Yangguang Li, Zi-Xin Zou, Zexiang Liu, Dehu Wang, Yuan Liang, Zhipeng Yu, Xingchao Liu, Yuan-Chen Guo, Ding Liang, Wanli Ouyang, et~al.
\newblock Triposg: High-fidelity 3d shape synthesis using large-scale rectified flow models.
\newblock \emph{arXiv preprint arXiv:2502.06608}, 2025{\natexlab{b}}.

\bibitem[Liang et~al.(2025{\natexlab{a}})Liang, Ma, Sun, Guo, and Zhang]{liang2025aligncvc}
Xinyue Liang, Zhiyuan Ma, Lingchen Sun, Yanjun Guo, and Lei Zhang.
\newblock Aligncvc: Aligning cross-view consistency for single-image-to-3d generation.
\newblock \emph{arXiv preprint arXiv:2506.23150}, 2025{\natexlab{a}}.

\bibitem[Liang et~al.(2025{\natexlab{b}})Liang, Luo, Chen, Chen, Yan, Li, Liu, and Tan]{liang2025unitex}
Yixun Liang, Kunming Luo, Xiao Chen, Rui Chen, Hongyu Yan, Weiyu Li, Jiarui Liu, and Ping Tan.
\newblock Unitex: Universal high fidelity generative texturing for 3d shapes.
\newblock \emph{arXiv preprint arXiv:2505.23253}, 2025{\natexlab{b}}.

\bibitem[Lin et~al.(2025)Lin, Pan, Yang, Li, and Mu]{lin2025diffsplat}
Chenguo Lin, Panwang Pan, Bangbang Yang, Zeming Li, and Yadong Mu.
\newblock Diffsplat: Repurposing image diffusion models for scalable gaussian splat generation.
\newblock \emph{arXiv preprint arXiv:2501.16764}, 2025.

\bibitem[Lin et~al.(2024)Lin, Liu, Li, and Yang]{lin2024common}
Shanchuan Lin, Bingchen Liu, Jiashi Li, and Xiao Yang.
\newblock Common diffusion noise schedules and sample steps are flawed.
\newblock In \emph{Proceedings of the IEEE/CVF winter conference on applications of computer vision}, pages 5404--5411, 2024.

\bibitem[Liu et~al.(2024)Liu, Wu, Liu, Liu, Wu, Peng, Zhao, Feng, Liu, and Ding]{liu2024texoct}
Jialun Liu, Chenming Wu, Xinqi Liu, Xing Liu, Jinbo Wu, Haotian Peng, Chen Zhao, Haocheng Feng, Jingtuo Liu, and Errui Ding.
\newblock Texoct: Generating textures of 3d models with octree-based diffusion.
\newblock In \emph{Proceedings of the IEEE/CVF Conference on Computer Vision and Pattern Recognition}, pages 4284--4293, 2024.

\bibitem[Liu et~al.(2023)Liu, Wu, Van~Hoorick, Tokmakov, Zakharov, and Vondrick]{liu2023zero}
Ruoshi Liu, Rundi Wu, Basile Van~Hoorick, Pavel Tokmakov, Sergey Zakharov, and Carl Vondrick.
\newblock Zero-1-to-3: Zero-shot one image to 3d object.
\newblock In \emph{Proceedings of the IEEE/CVF international conference on computer vision}, pages 9298--9309, 2023.

\bibitem[Long et~al.(2024)Long, Guo, Lin, Liu, Dou, Liu, Ma, Zhang, Habermann, Theobalt, et~al.]{long2024wonder3d}
Xiaoxiao Long, Yuan-Chen Guo, Cheng Lin, Yuan Liu, Zhiyang Dou, Lingjie Liu, Yuexin Ma, Song-Hai Zhang, Marc Habermann, Christian Theobalt, et~al.
\newblock Wonder3d: Single image to 3d using cross-domain diffusion.
\newblock In \emph{Proceedings of the IEEE/CVF conference on computer vision and pattern recognition}, pages 9970--9980, 2024.

\bibitem[Loshchilov and Hutter(2017)]{loshchilov2017decoupled}
Ilya Loshchilov and Frank Hutter.
\newblock Decoupled weight decay regularization.
\newblock \emph{arXiv preprint arXiv:1711.05101}, 2017.

\bibitem[Ma et~al.(2024)Ma, Wei, Zhang, Zhu, Lei, and Zhang]{ma2024scaledreamer}
Zhiyuan Ma, Yuxiang Wei, Yabin Zhang, Xiangyu Zhu, Zhen Lei, and Lei Zhang.
\newblock Scaledreamer: Scalable text-to-3d synthesis with asynchronous score distillation.
\newblock In \emph{European Conference on Computer Vision}, pages 1--19. Springer, 2024.

\bibitem[Ma et~al.(2025)Ma, Liang, Wu, Zhu, Lei, and Zhang]{ma2025progressive}
Zhiyuan Ma, Xinyue Liang, Rongyuan Wu, Xiangyu Zhu, Zhen Lei, and Lei Zhang.
\newblock Progressive rendering distillation: Adapting stable diffusion for instant text-to-mesh generation without 3d data.
\newblock In \emph{Proceedings of the Computer Vision and Pattern Recognition Conference}, pages 11036--11050, 2025.

\bibitem[Melas-Kyriazi et~al.(2023)Melas-Kyriazi, Laina, Rupprecht, and Vedaldi]{melas2023realfusion}
Luke Melas-Kyriazi, Iro Laina, Christian Rupprecht, and Andrea Vedaldi.
\newblock Realfusion: 360deg reconstruction of any object from a single image.
\newblock In \emph{Proceedings of the IEEE/CVF conference on computer vision and pattern recognition}, pages 8446--8455, 2023.

\bibitem[Poole et~al.(2022)Poole, Jain, Barron, and Mildenhall]{poole2022dreamfusion}
Ben Poole, Ajay Jain, Jonathan~T Barron, and Ben Mildenhall.
\newblock Dreamfusion: Text-to-3d using 2d diffusion.
\newblock \emph{arXiv preprint arXiv:2209.14988}, 2022.

\bibitem[Qiu et~al.(2024)Qiu, Chen, Gu, Zuo, Xu, Wu, Yuan, Dong, Bo, and Han]{qiu2024richdreamer}
Lingteng Qiu, Guanying Chen, Xiaodong Gu, Qi Zuo, Mutian Xu, Yushuang Wu, Weihao Yuan, Zilong Dong, Liefeng Bo, and Xiaoguang Han.
\newblock Richdreamer: A generalizable normal-depth diffusion model for detail richness in text-to-3d.
\newblock In \emph{Proceedings of the IEEE/CVF conference on computer vision and pattern recognition}, pages 9914--9925, 2024.

\bibitem[Radford et~al.(2021)Radford, Kim, Hallacy, Ramesh, Goh, Agarwal, Sastry, Askell, Mishkin, Clark, et~al.]{radford2021learning}
Alec Radford, Jong~Wook Kim, Chris Hallacy, Aditya Ramesh, Gabriel Goh, Sandhini Agarwal, Girish Sastry, Amanda Askell, Pamela Mishkin, Jack Clark, et~al.
\newblock Learning transferable visual models from natural language supervision.
\newblock In \emph{International conference on machine learning}, pages 8748--8763. PmLR, 2021.

\bibitem[Reizenstein et~al.(2021)Reizenstein, Shapovalov, Henzler, Sbordone, Labatut, and Novotny]{reizenstein21co3d}
Jeremy Reizenstein, Roman Shapovalov, Philipp Henzler, Luca Sbordone, Patrick Labatut, and David Novotny.
\newblock Common objects in 3d: Large-scale learning and evaluation of real-life 3d category reconstruction.
\newblock In \emph{International Conference on Computer Vision}, 2021.

\bibitem[Russakovsky et~al.(2015)Russakovsky, Deng, Su, Krause, Satheesh, Ma, Huang, Karpathy, Khosla, Bernstein, Berg, and Fei-Fei]{imagenet15russakovsky}
Olga Russakovsky, Jia Deng, Hao Su, Jonathan Krause, Sanjeev Satheesh, Sean Ma, Zhiheng Huang, Andrej Karpathy, Aditya Khosla, Michael Bernstein, Alexander~C. Berg, and Li Fei-Fei.
\newblock {ImageNet Large Scale Visual Recognition Challenge}.
\newblock \emph{International Journal of Computer Vision (IJCV)}, 115\penalty0 (3):\penalty0 211--252, 2015.

\bibitem[Schuhmann(2022)]{improved_aesthetic_predictor}
Christoph Schuhmann.
\newblock Improved aesthetic predictor.
\newblock \url{https://github.com/christophschuhmann/improved-aesthetic-predictor}, 2022.
\newblock Accessed: 2025-11-05.

\bibitem[Siddiqui et~al.(2024)Siddiqui, Alliegro, Artemov, Tommasi, Sirigatti, Rosov, Dai, and Nie{\ss}ner]{siddiqui2024meshgpt}
Yawar Siddiqui, Antonio Alliegro, Alexey Artemov, Tatiana Tommasi, Daniele Sirigatti, Vladislav Rosov, Angela Dai, and Matthias Nie{\ss}ner.
\newblock Meshgpt: Generating triangle meshes with decoder-only transformers.
\newblock In \emph{Proceedings of the IEEE/CVF conference on computer vision and pattern recognition}, pages 19615--19625, 2024.

\bibitem[Sim{\'e}oni et~al.(2025)Sim{\'e}oni, Vo, Seitzer, Baldassarre, Oquab, Jose, Khalidov, Szafraniec, Yi, Ramamonjisoa, et~al.]{simeoni2025dinov3}
Oriane Sim{\'e}oni, Huy~V Vo, Maximilian Seitzer, Federico Baldassarre, Maxime Oquab, Cijo Jose, Vasil Khalidov, Marc Szafraniec, Seungeun Yi, Micha{\"e}l Ramamonjisoa, et~al.
\newblock Dinov3.
\newblock \emph{arXiv preprint arXiv:2508.10104}, 2025.

\bibitem[Talebi and Milanfar(2018)]{talebi2018nima}
Hossein Talebi and Peyman Milanfar.
\newblock Nima: Neural image assessment.
\newblock \emph{IEEE transactions on image processing}, 27\penalty0 (8):\penalty0 3998--4011, 2018.

\bibitem[Tang et~al.(2023)Tang, Wang, Zhang, Zhang, Yi, Ma, and Chen]{tang2023make}
Junshu Tang, Tengfei Wang, Bo Zhang, Ting Zhang, Ran Yi, Lizhuang Ma, and Dong Chen.
\newblock Make-it-3d: High-fidelity 3d creation from a single image with diffusion prior.
\newblock In \emph{Proceedings of the IEEE/CVF international conference on computer vision}, pages 22819--22829, 2023.

\bibitem[Tang et~al.(2024)Tang, Chen, Chen, Wang, Zeng, and Liu]{tang2024lgm}
Jiaxiang Tang, Zhaoxi Chen, Xiaokang Chen, Tengfei Wang, Gang Zeng, and Ziwei Liu.
\newblock Lgm: Large multi-view gaussian model for high-resolution 3d content creation.
\newblock In \emph{European Conference on Computer Vision}, pages 1--18. Springer, 2024.

\bibitem[Wang et~al.(2024{\natexlab{a}})Wang, Wang, Li, Zhang, Lei, and Zhang]{wang2024open}
Pengfei Wang, Yuxi Wang, Shuai Li, Zhaoxiang Zhang, Zhen Lei, and Lei Zhang.
\newblock Open vocabulary 3d scene understanding via geometry guided self-distillation.
\newblock In \emph{European Conference on Computer Vision}, pages 442--460. Springer, 2024{\natexlab{a}}.

\bibitem[Wang et~al.(2026)Wang, Chen, Ma, Guo, Zhang, and Zhang]{wang2026one2scene}
Pengfei Wang, Liyi Chen, Zhiyuan Ma, Yanjun Guo, Guowen Zhang, and Lei Zhang.
\newblock One2scene: Geometric consistent explorable 3d scene generation from a single image.
\newblock \emph{arXiv preprint arXiv:2602.19766}, 2026.

\bibitem[Wang et~al.(2023)Wang, Lu, Wang, Bao, Li, Su, and Zhu]{wang2023prolificdreamer}
Zhengyi Wang, Cheng Lu, Yikai Wang, Fan Bao, Chongxuan Li, Hang Su, and Jun Zhu.
\newblock Prolificdreamer: High-fidelity and diverse text-to-3d generation with variational score distillation.
\newblock \emph{Advances in neural information processing systems}, 36:\penalty0 8406--8441, 2023.

\bibitem[Wang et~al.(2024{\natexlab{b}})Wang, Lorraine, Wang, Su, Zhu, Fidler, and Zeng]{wang2024llama}
Zhengyi Wang, Jonathan Lorraine, Yikai Wang, Hang Su, Jun Zhu, Sanja Fidler, and Xiaohui Zeng.
\newblock Llama-mesh: Unifying 3d mesh generation with language models.
\newblock \emph{arXiv preprint arXiv:2411.09595}, 2024{\natexlab{b}}.

\bibitem[Wang et~al.(2022)Wang, Montoya, Munechika, Yang, Hoover, and Chau]{wang2022diffusiondb}
Zijie~J Wang, Evan Montoya, David Munechika, Haoyang Yang, Benjamin Hoover, and Duen~Horng Chau.
\newblock Diffusiondb: A large-scale prompt gallery dataset for text-to-image generative models.
\newblock \emph{arXiv preprint arXiv:2210.14896}, 2022.

\bibitem[Xiang et~al.(2025)Xiang, Lv, Xu, Deng, Wang, Zhang, Chen, Tong, and Yang]{xiang2025structured}
Jianfeng Xiang, Zelong Lv, Sicheng Xu, Yu Deng, Ruicheng Wang, Bowen Zhang, Dong Chen, Xin Tong, and Jiaolong Yang.
\newblock Structured 3d latents for scalable and versatile 3d generation.
\newblock In \emph{Proceedings of the Computer Vision and Pattern Recognition Conference}, pages 21469--21480, 2025.

\bibitem[Xiong et~al.(2025)Xiong, Liu, Hu, Wu, Wu, Liu, Zhao, Ding, and Lian]{xiong2025texgaussian}
Bojun Xiong, Jialun Liu, Jiakui Hu, Chenming Wu, Jinbo Wu, Xing Liu, Chen Zhao, Errui Ding, and Zhouhui Lian.
\newblock Texgaussian: Generating high-quality pbr material via octree-based 3d gaussian splatting.
\newblock In \emph{Proceedings of the Computer Vision and Pattern Recognition Conference}, pages 551--561, 2025.

\bibitem[Yang et~al.(2022)Yang, Wu, Shi, Lao, Gong, Cao, Wang, and Yang]{yang2022maniqa}
Sidi Yang, Tianhe Wu, Shuwei Shi, Shanshan Lao, Yuan Gong, Mingdeng Cao, Jiahao Wang, and Yujiu Yang.
\newblock Maniqa: Multi-dimension attention network for no-reference image quality assessment.
\newblock In \emph{Proceedings of the IEEE/CVF conference on computer vision and pattern recognition}, pages 1191--1200, 2022.

\bibitem[Ye et~al.(2025)Ye, Wu, Lu, Chang, Guo, Zhou, Zhao, and Han]{ye2025hi3dgen}
Chongjie Ye, Yushuang Wu, Ziteng Lu, Jiahao Chang, Xiaoyang Guo, Jiaqing Zhou, Hao Zhao, and Xiaoguang Han.
\newblock Hi3dgen: High-fidelity 3d geometry generation from images via normal bridging.
\newblock \emph{arXiv preprint arXiv:2503.22236}, 3:\penalty0 2, 2025.

\bibitem[Yu et~al.(2024)Yu, Yuan, Guo, Liu, Liu, Li, Cao, Liang, and Qi]{yu2024texgen}
Xin Yu, Ze Yuan, Yuan-Chen Guo, Ying-Tian Liu, Jianhui Liu, Yangguang Li, Yan-Pei Cao, Ding Liang, and Xiaojuan Qi.
\newblock Texgen: a generative diffusion model for mesh textures.
\newblock \emph{ACM Transactions on Graphics (TOG)}, 43\penalty0 (6):\penalty0 1--14, 2024.

\bibitem[Yushi et~al.(2025)Yushi, Zhou, Lyu, Hong, Yang, Dai, Pan, and Loy]{yushi2025gaussiananything}
LAN Yushi, Shangchen Zhou, Zhaoyang Lyu, Fangzhou Hong, Shuai Yang, Bo Dai, Xingang Pan, and Chen~Change Loy.
\newblock Gaussiananything: Interactive point cloud flow matching for 3d generation.
\newblock In \emph{The Thirteenth International Conference on Learning Representations}, 2025.

\bibitem[Zhang et~al.(2025)Zhang, Zhang, Jiang, Bai, Yang, Xu, and Yu]{zhang2025bang}
Longwen Zhang, Qixuan Zhang, Haoran Jiang, Yinuo Bai, Wei Yang, Lan Xu, and Jingyi Yu.
\newblock Bang: Dividing 3d assets via generative exploded dynamics.
\newblock \emph{ACM Transactions on Graphics (TOG)}, 44\penalty0 (4):\penalty0 1--21, 2025.

\bibitem[Zhao et~al.(2025{\natexlab{a}})Zhao, Ye, Wang, Liu, Chen, Wang, and Zhu]{zhao2025deepmesh}
Ruowen Zhao, Junliang Ye, Zhengyi Wang, Guangce Liu, Yiwen Chen, Yikai Wang, and Jun Zhu.
\newblock Deepmesh: Auto-regressive artist-mesh creation with reinforcement learning.
\newblock In \emph{Proceedings of the IEEE/CVF International Conference on Computer Vision}, pages 10612--10623, 2025{\natexlab{a}}.

\bibitem[Zhao et~al.(2025{\natexlab{b}})Zhao, Lai, Lin, Zhao, Liu, Yang, Feng, Yang, Zhang, Yang, et~al.]{zhao2025hunyuan3d}
Zibo Zhao, Zeqiang Lai, Qingxiang Lin, Yunfei Zhao, Haolin Liu, Shuhui Yang, Yifei Feng, Mingxin Yang, Sheng Zhang, Xianghui Yang, et~al.
\newblock Hunyuan3d 2.0: Scaling diffusion models for high resolution textured 3d assets generation.
\newblock \emph{arXiv preprint arXiv:2501.12202}, 2025{\natexlab{b}}.

\end{thebibliography}
}

% WARNING: do not forget to delete the supplementary pages from your submission 
\clearpage
\appendix
\setcounter{page}{1}

\twocolumn[
\begin{center}
\Large \textbf{Appendix}
\end{center}
\vspace{4em}
]

\section{More Details for Photo3D-MV}
\label{sec:Photo3D-MV}
As mentioned in the main paper, we developed a structure-aligned multi-view synthesis pipeline to construct the detail-enhanced dataset Photo3D-MV. In this pipeline, we first process the text prompts, then generate the corresponding images, construct 3D models from these images, and finally refine the rendered results to achieve greater realism.

Specifically, for the text prompt processing stage, we employ LLaMA-3-8B~\cite{llama3modelcard} to transform the input text from  DiffusionDB~\cite{wang2022diffusiondb} into object-centered descriptions, while appending unified realistic constraints to ensure photorealistic image generation.
For the final realistic multi-view generation stage, we use GPT-4o-Image to generate detail-enhanced multi-views on the rendered images with an editing prompt.
The textual prompts we used are as follows:

\begin{tcolorbox}[
  colback=gray!2,
  colframe=gray!25,
  title=\textbf{Text Prompt Processing (Input for LLaMA-3-8B)},
  coltitle=black, % <- 标题文字为黑色
  boxrule=0.4pt,
  arc=2pt,
  fonttitle=\bfseries,
  left=4pt,right=4pt,top=3pt,bottom=3pt
]
\small
``Optimize this prompt into a single, high-quality, photorealistic physical object description, 
focusing on realistic materials, detailed textures, and authentic visual qualities:
\textit{\{Raw\_Text\}}.''
\end{tcolorbox}

\begin{tcolorbox}[
  colback=gray!2,
  colframe=gray!25,
  title=\textbf{Realistic Constraints  (Input for Flux.1‑Dev~\cite{flux2024}}),
  coltitle=black, % <- 同样这里
  boxrule=0.4pt,
  arc=2pt,
  left=4pt,right=4pt,top=3pt,bottom=3pt
]
\small
``\textit{\{Text\_Prompt\}}, real camera shot, real photograph, pure white background with no shadows, complete object, high-quality photography, macro lens detail, professional studio lighting.''
\end{tcolorbox}

\begin{tcolorbox}[
  colback=gray!2,
  colframe=gray!25,
  title=\textbf{Realistic Multi-View Generation (Input for GPT-4o-Image~\cite{hurst2024gpt}}),
  coltitle=black, % <- 保持黑色标题
  boxrule=0.4pt,
  arc=2pt,
  left=4pt,right=4pt,top=3pt,bottom=3pt
]
\small
``Edit Image, photorealistic micro-refinement only, make it a real object; strictly preserve exact composition and framing (NO recomposition); 
lock camera parameters (position, rotation, FOV, focal length); lock scale and subject position; preserve exact geometry, silhouette and perspective; 
fix tiny artifacts; refine textures and micro-details; keep colors and lighting exactly the same.''
\end{tcolorbox}
\begin{figure*}[h]
    \centering
    \includegraphics[width=0.7\linewidth]{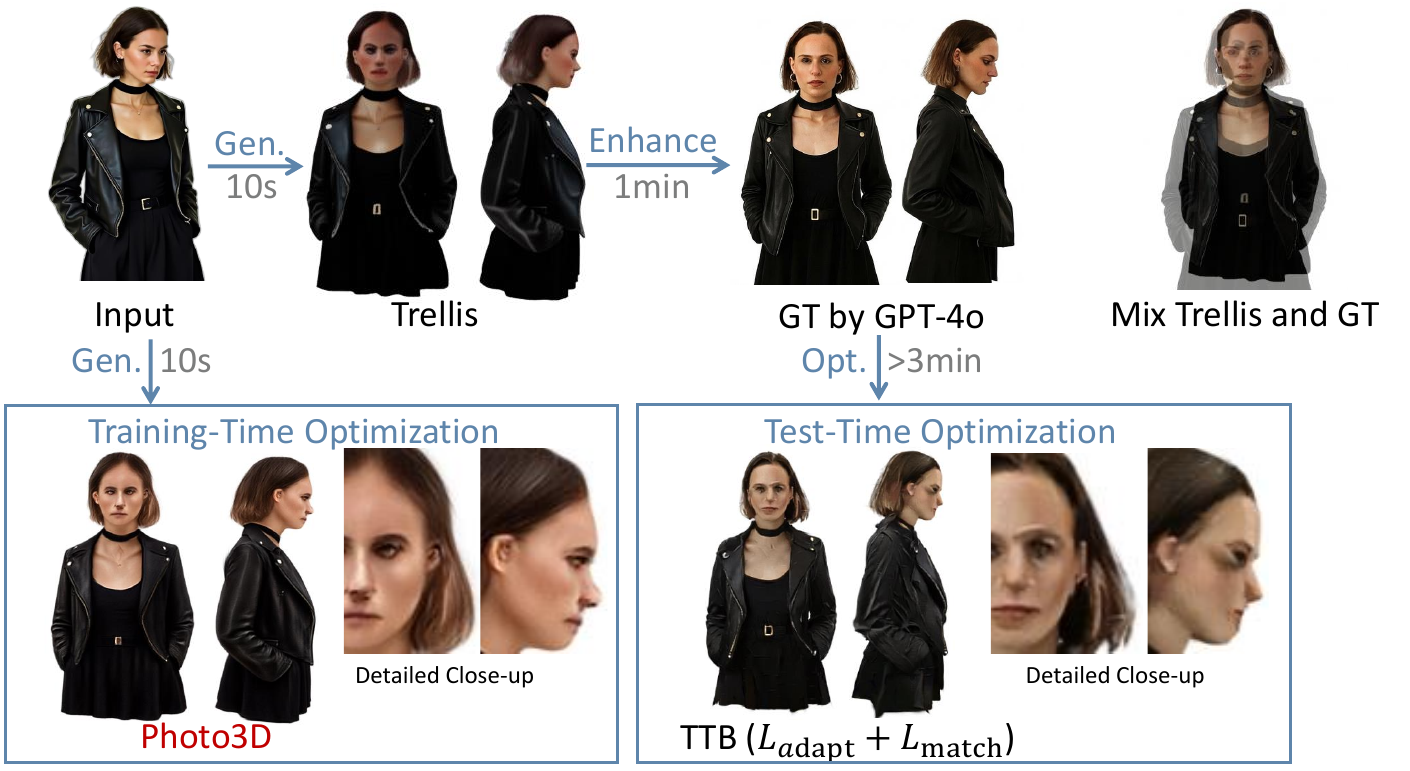}
    \caption{Comparison between the test-time optimization baseline and Photo3D (training-time optimization).}
    \label{fig:cvc}
\end{figure*}
\begin{figure*}[h]
    \centering
    \includegraphics[width=0.5\linewidth]{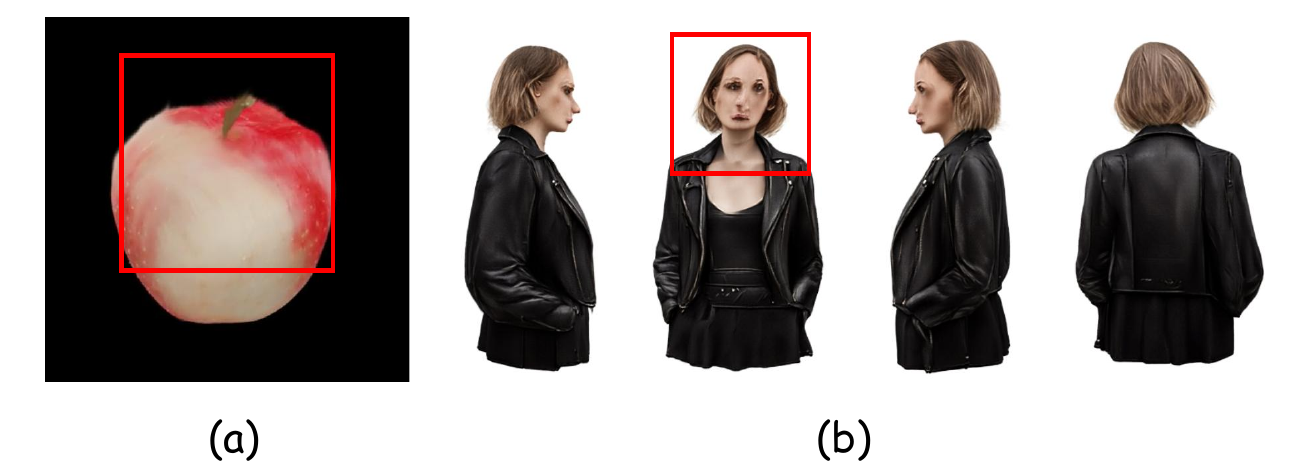}
    \caption{(a) Incomplete view coverage causes blurred regions. (b) Inconsistent details across views lead to distorted 3D structures.}
    \label{fig:fullcover}
\end{figure*}
 \begin{figure*}[!htbp]
     \centering
     \includegraphics[width=0.9\linewidth]{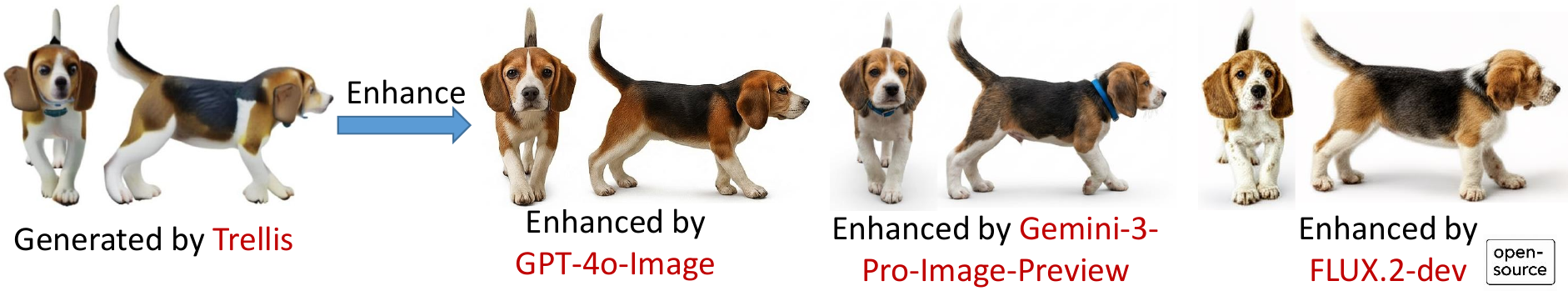}
     \caption{Enhanced multi-views of different 2D generators.}
     \label{fig:choices}
 \end{figure*}

Besides, we also evaluated several recent advanced image generators for realistic multi-view generation, 
including GPT‑4o‑Image, Gemini‑2.5‑Flash~\cite{comanici2025gemini}, and Flux.1‑Kontext~\cite{labs2025flux1kontextflowmatching}, which are capable of both text and image conditioned generation. 
Notably, GPT‑4o‑Image achieves superior performance in fine‑grained realistic detail enhancement 
and is therefore adopted in our framework. 
Since the Trellis-generated 3D models will have some color distortions, we further align the generated views with the rendered views by performing per‑channel histogram matching in the CIE L*a*b*~\cite{dmenuaclcommission} color space, so that the luminance and color distributions can match the original rendered views while preserving the 3D structure.
%Comparisons among multi-views generated by different image generators are shown in Fig.~\ref{fig:generator}.
%\begin{figure*}[h]
 %   \centering
  %  \includegraphics[width=0.9\linewidth]{figs/generator.pdf}
   % \caption{Comparisons of realistic-detail-enhanced multi-view generation with different image generators.}
   % \label{fig:generator}
%\end{figure*}

Compared to original Trellis outputs, after GPT-4o-Image enhancement, multi-view consistency of Photo3D-MV defined by Probe3D~\cite{elbanani2024probing} slightly changes from 0.854 to 0.842 \textcolor{Red}{(1\%$\downarrow$)}, while realism measured by MANIQA~\cite{yang2022maniqa} substantially 
 improves from 0.438 to 0.655 \textcolor{ForestGreen}{(\textbf{50\%}$\uparrow$)}, indicating that the enhancement boosts realism without severe consistency degradation.

\section{Analyses on Training Strategies}
\label{sec:analyse}

\subsection{Test-Time Optimization Baseline}
We show the test-time optimization results in Fig.~\ref{fig:cvc} to explain our proposal of training the 3D generators instead of using a 2D generator for multi-view test-time enhancement. The results highlight that improving the 3D model itself leads to more consistent and geometry-aware realism.
Compared to Photo3D, test-time optimization baseline significantly degrades both quality (MUSIQ: 76.6$\rightarrow$ 71.5) and time efficiency (10\,s$\rightarrow$ $>$4\,min).
This is mainly caused by the mismatches with the initial 3D geometry (see Fig.~\ref{fig:cvc} \textbf{Mix Trellis and GT}) in the enhanced views,
which inevitably lead to texture distortions during test-time optimization, whereas Photo3D aligns realistic details with the 3D-native distribution, 
effectively preserving geometric consistency.

\subsection{ Training-Time Optimization Strategies}

We analyze different training strategies for diffusion-based geometry–texture coupled 3D generation. 
In the original training paradigm of the geometry–texture coupled 3D-native generator Trellis, each 3D asset is first rendered into about 150 multi-view images, which are then projected onto the asset’s voxelized representation and subsequently encoded into a structured 3D latent.
Following the original Trellis paradigm, we attempt to construct GT 3D latents by projecting the 4 realism‑enhanced multi‑view images onto the sparse structures generated in the preceding 3D sparse structure generation stage of Trellis, 
and jointly train the 3D VAE models on the resulting structured 3D latents under the supervision of realism‑enhanced multi‑view images.
However, since the reconstructed structured 3D latent appearance mainly depends on the information provided during projection, four orthogonal views cannot fully cover the latent volume, leading to blurred regions in the unseen areas (Fig.~\ref{fig:fullcover}(a)), or ensure detail consistency across views (Fig.~\ref{fig:fullcover}(b)), resulting in distorted 3D structures.
Consequently, such structured GT 3D latents provide insufficient supervision for realistic 3D generation. Our proposed realism-enhancement scheme offers a more robust solution that overcomes this limitation.

\section{Choices of 2D generators}
\label{sec:choice}
Our framework is not sensitive to specific 2D generators and can work with any high‑quality image model.  More recent models, such as Gemini‑3‑Pro‑Image‑Preview~\cite{gemini3pro} and the open-source 
model FLUX.2‑dev~\cite{flux-2-2025}, also work well with our proposed realistic multi-view synthesis pipeline, as shown in Fig.~\ref{fig:choices}.

\section{More Results}

\label{sec:morere}
\begin{figure*}
    \centering
    \includegraphics[width=1\linewidth]{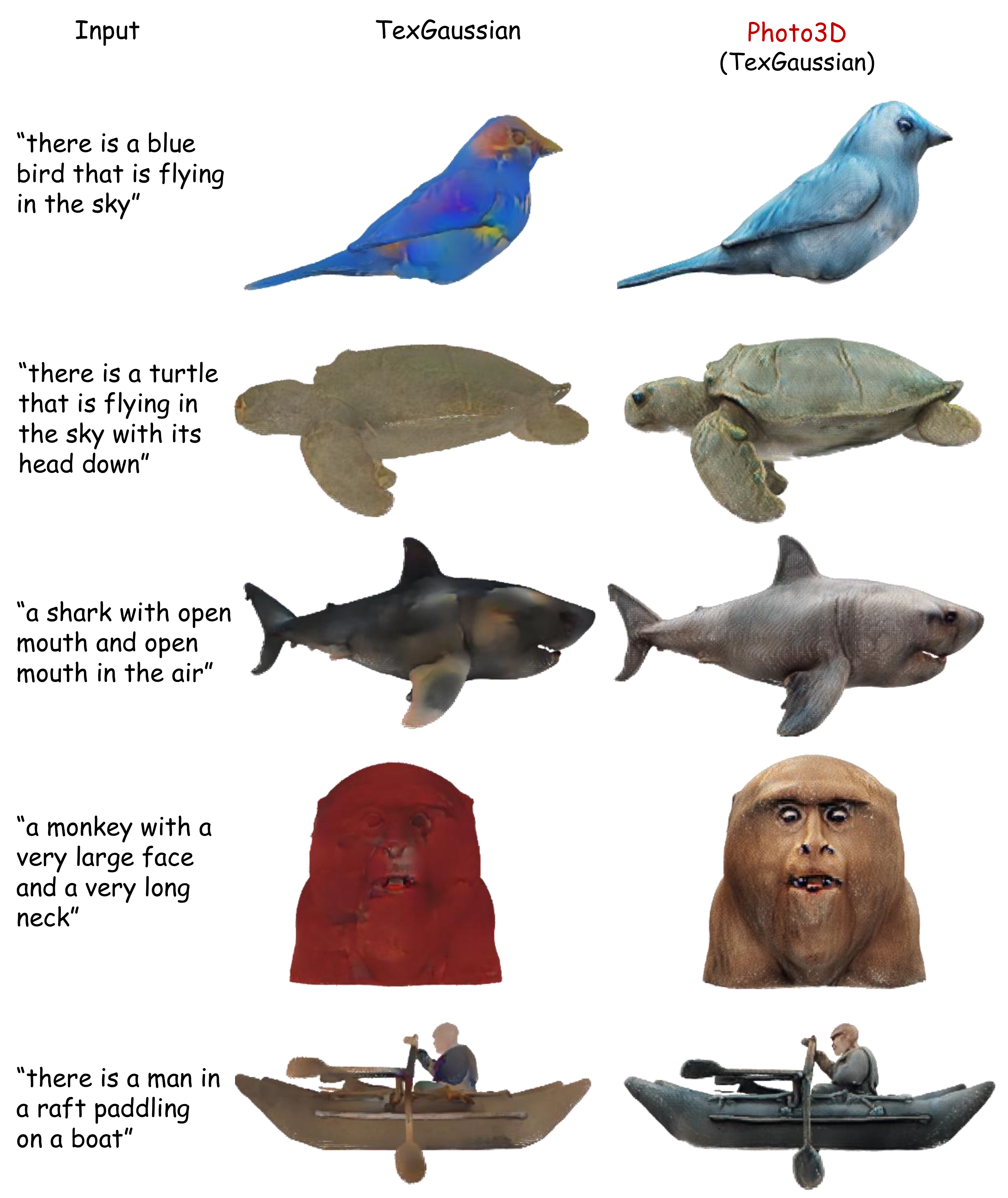}
     \caption{Comparison of 3D generation results between TexGaussian~\cite{xiong2025texgaussian} and Photo3D trained on the TexGaussian's 3D-native texturing model. The results demonstrate the realistic appearance of geometry-texture decoupled 3D-native generation achieved by Photo3D.}
    \label{fig:texgaussian}
\end{figure*}

\begin{figure*}
    \centering
    \includegraphics[width=1\linewidth]{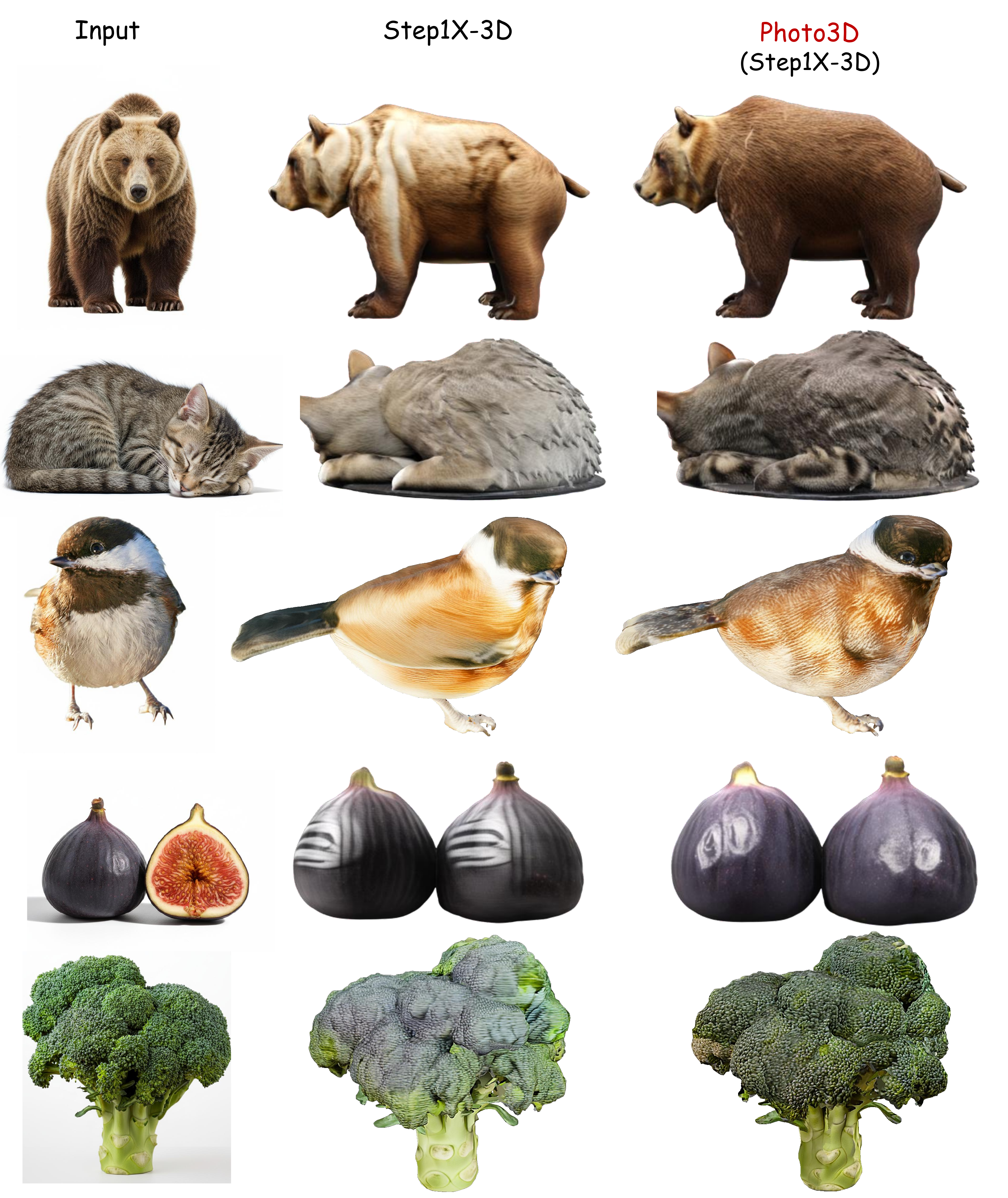}
    \caption{Comparison of 3D generation results between Step1X-3D~\cite{li2025step1x} and Photo3D trained on the Step1X-3D's multi-view texturing model. The results demonstrate the realistic appearance of geometry-texture decoupled 3D-native generation achieved by Photo3D.}
    \label{fig:step}
\end{figure*}

\begin{figure*}
    \centering
    \includegraphics[width=0.99\linewidth]{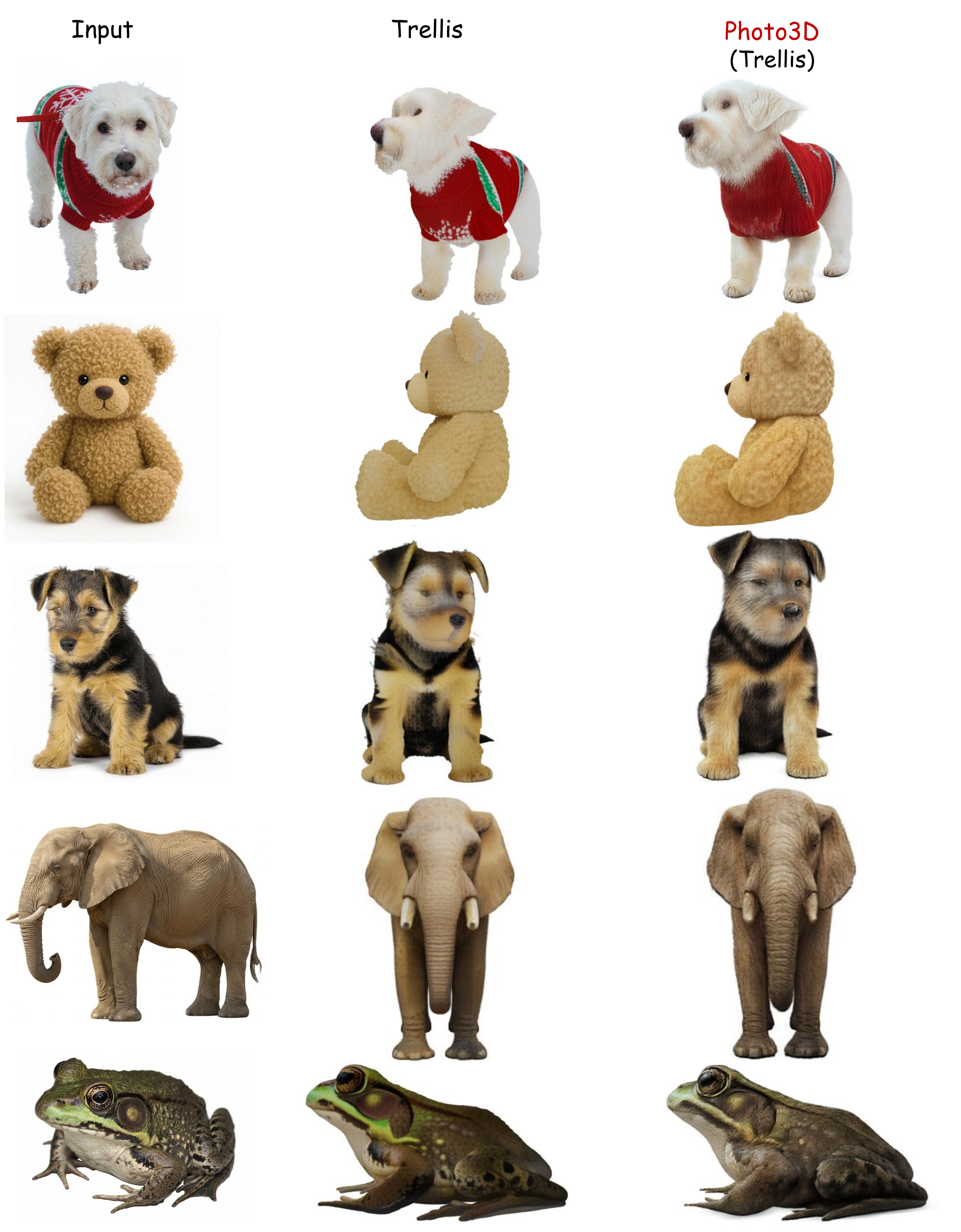}
    \caption{Comparison of 3D generation results between Trellis~\cite{xiang2025structured} and Photo3D trained on the Trellis model. The results demonstrate the realistic appearance of geometry-texture coupled 3D-native generation achieved by Photo3D.}
    \label{fig:trellis}
\end{figure*}

We present more 3D generation results produced by Photo3D, together with those of its baseline counterparts, in Fig.\ref{fig:texgaussian}, Fig.~\ref{fig:step} and Fig.\ref{fig:trellis}.
Note that for the 3D-native texturing model TexGaussian~\cite{xiong2025texgaussian}, we first input the images into Step1X-3D to obtain untextured 3D meshes. Subsequently, we use BLIP-2~\cite{li2023blip} to generate captions for each image, which are then adopted as text conditions for both TexGaussian and Photo3D (TexGaussian).

\section{Geometric Correction}

\label{sec:corr}

\begin{figure}[h]
    \centering
    \includegraphics[width=1\linewidth]{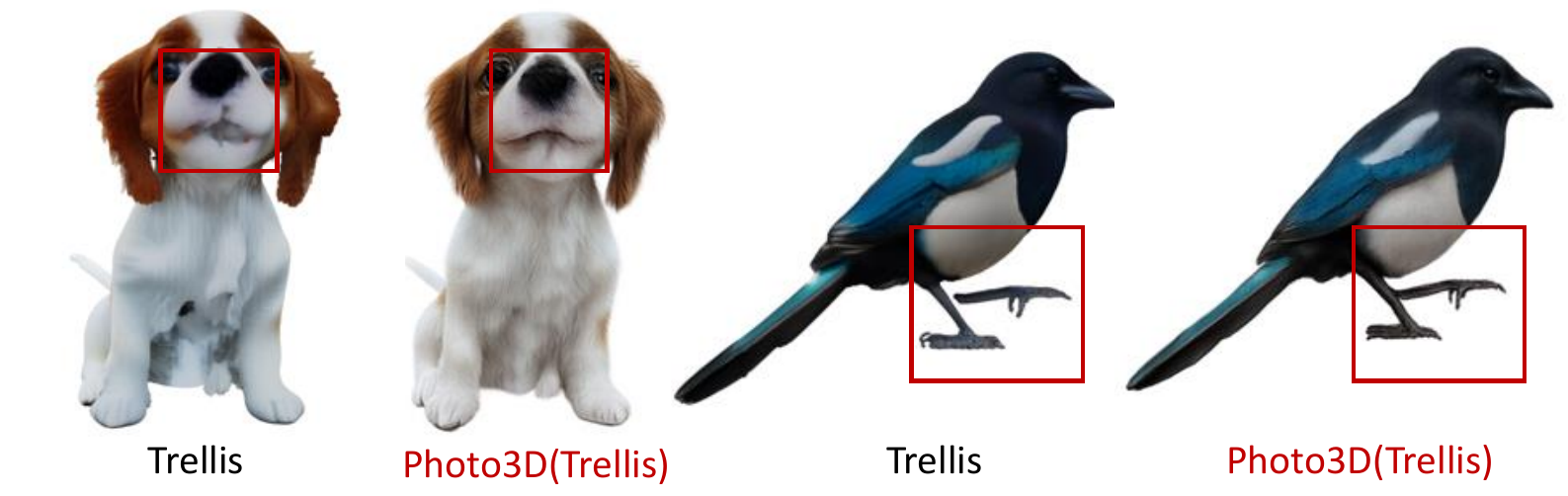}
    \caption{Geometric correction ability of Photo3D (Trellis).}
    \label{fig:gc}
\end{figure}

For Photo3D (Trellis), our method can correct geometry flaws (see examples in Fig.~\ref{fig:gc}). This is achieved via the learned realistic appearance priors under coupled geometry–texture optimization. For geometry–texture decoupled methods (TexGaussian and Step1X-3D), the geometry is fixed in the texturing stage and thus requires a good base geometry generated in the previous stage.

\label{sec:results}

%We further discuss the supervision designs for multi-view texturing models.

\end{document}